%% file: 000_main.tex
\documentclass{article}

\usepackage[preprint]{neurips_2025}
\usepackage{booktabs}
\usepackage{enumitem}
\usepackage[table]{xcolor}
\usepackage{multirow}
\usepackage{wrapfig}
\usepackage{xspace}
\usepackage{dsfont}
\usepackage{amssymb}
\usepackage{soul}
\usepackage{fancyvrb}
\usepackage{colortbl}
\usepackage{subcaption}

\setcitestyle{numbers,square}

\usepackage[utf8]{inputenc} %
\usepackage[T1]{fontenc}    %
\usepackage{hyperref}       %
\usepackage{url}   
\usepackage{stmaryrd}
\usepackage{booktabs}       %
\usepackage{amsfonts}       %
\usepackage{nicefrac}       %
\usepackage{microtype}      %
\usepackage{xcolor}         %
\usepackage{algorithm}
\usepackage[noend]{algpseudocode}
\algrenewcommand\algorithmicindent{1.0em}%
\usepackage[textsize=scriptsize]{todonotes}
\usepackage{tikz}

\definecolor{nicegray}{HTML}{9CA3AF} %
\definecolor{utorange}{HTML}{BF5700}
\definecolor{mendblue}{HTML}{00A2FF}

\newcommand{\mydagger}[0]{\textsuperscript{\textdagger}}
\usepackage[skins,breakable]{tcolorbox}
\definecolor{light-purple}{RGB}{151,156,171}
\definecolor{blue-color}{RGB}{40,166,189}
\definecolor{pink-color}{RGB}{237,46,104} 
\definecolor{dark-grey-color}{RGB}{79,91,102}

\newcommand*\circled[1]{\tikz[baseline=(char.base)]{
            \node[shape=circle,draw,inner sep=0.8pt] (char) {#1};}}
            
\newtcolorbox[list inside=prompt,auto counter,number within=section]{prompt}[1][]{
    colbacktitle=black!80,
    colframe=black!80,
    coltitle=white,
    fonttitle=\sffamily\bfseries,
    fontupper=\ttfamily,
    boxsep=5pt,
    left=0pt,
    right=0pt,
    top=0pt,
    bottom=0pt,
    boxrule=1pt,
    enhanced, 
    breakable,
    skin first=enhanced,
    skin middle=enhanced,
    skin last=enhanced,
    #1,
}

\usepackage{fontawesome5}
\usepackage{hyperref}

\makeatletter
\newcommand{\github}[1]{%
   \href{#1}{\faGithubSquare}%
}
\makeatother

\newcommand{\catchyname}[0]{\texttt{PropMEND}\xspace}
\newcommand{\realdata}[0]{\texttt{RippleEdit}\xspace}
\newcommand{\syndata}[0]{\texttt{Controlled RippleEdit}\xspace}
\newcommand{\syndataSmall}[0]{\texttt{Controlled RippleEdit}\xspace}
\newcommand\resulttablefontsize{\fontsize{7.7pt}{9.24pt}\selectfont}
\usepackage{amsmath,amssymb}
\DeclareMathOperator{\E}{\mathbb{E}}

\title{\catchyname: Hypernetworks for \\ Knowledge Propagation in LLMs}
\author{
Zeyu Leo Liu$^{\dagger}$\;\;\;\;
Greg Durrett$^{\dagger}$\;\;\;\;
Eunsol Choi$^{\ddagger}$\\
$^{\dagger}$ The University of Texas at Austin\;\;\;\;$^{\ddagger}$ New York University \\
\url{zliu@cs.utexas.edu}\\
}

\input{commands}

\begin{document}

\maketitle

\begin{abstract}
Knowledge editing techniques for large language models (LLMs) can inject knowledge that is later reproducible verbatim, but they fall short on \emph{propagating} that knowledge: models cannot answer questions that require reasoning with the injected knowledge. We present a hypernetwork-based approach for knowledge propagation, named \catchyname{}, where we meta-learn how to modify gradients of a language modeling loss to encourage injected information to propagate. Our approach extends the meta-objective of MEND \cite{mend} so that gradient updates on knowledge are transformed to enable answering multi-hop questions involving that knowledge. We show improved performance on the \realdata dataset, showing almost $2\times$ accuracy on challenging multi-hop questions whose answers are not explicitly stated in the injected fact. We further introduce a new dataset, \syndata, to evaluate the generalization of our hypernetwork, testing knowledge propagation along relations and entities unseen during hypernetwork training. \catchyname still outperforms existing approaches in unseen entity-relation pairs, yet the performance gap decreases substantially, suggesting future work in propagating knowledge to a wide range of relations.

\end{abstract}

\section{Introduction}%

Knowledge editing methods~\cite{rome, mend, Nicola_De_Cao_21_KE, Sinitsin2020Editable} can transform large language models (LLMs) to \emph{reproduce} injected knowledge, but induce very limited \emph{propagation} of that knowledge~\cite{ripple_edit,zhong2025mquakeremastered}. This failure stands in disappointing contrast to LLMs' ability to propagate knowledge that is given in context at inference time \cite{Onoe2022EntityCB,IKE}. One promising path for propagation is through training on data that explicitly demonstrates that propagation \cite{prop_by_distill,DCT,minjoon_fact}, but these methods require large-scale data augmentation for each piece of knowledge to be injected \cite{synCPT}.%

\begin{figure}[t!]
  \centering
  \includegraphics[width=\textwidth,trim={5em 49em 10em 2em},clip]{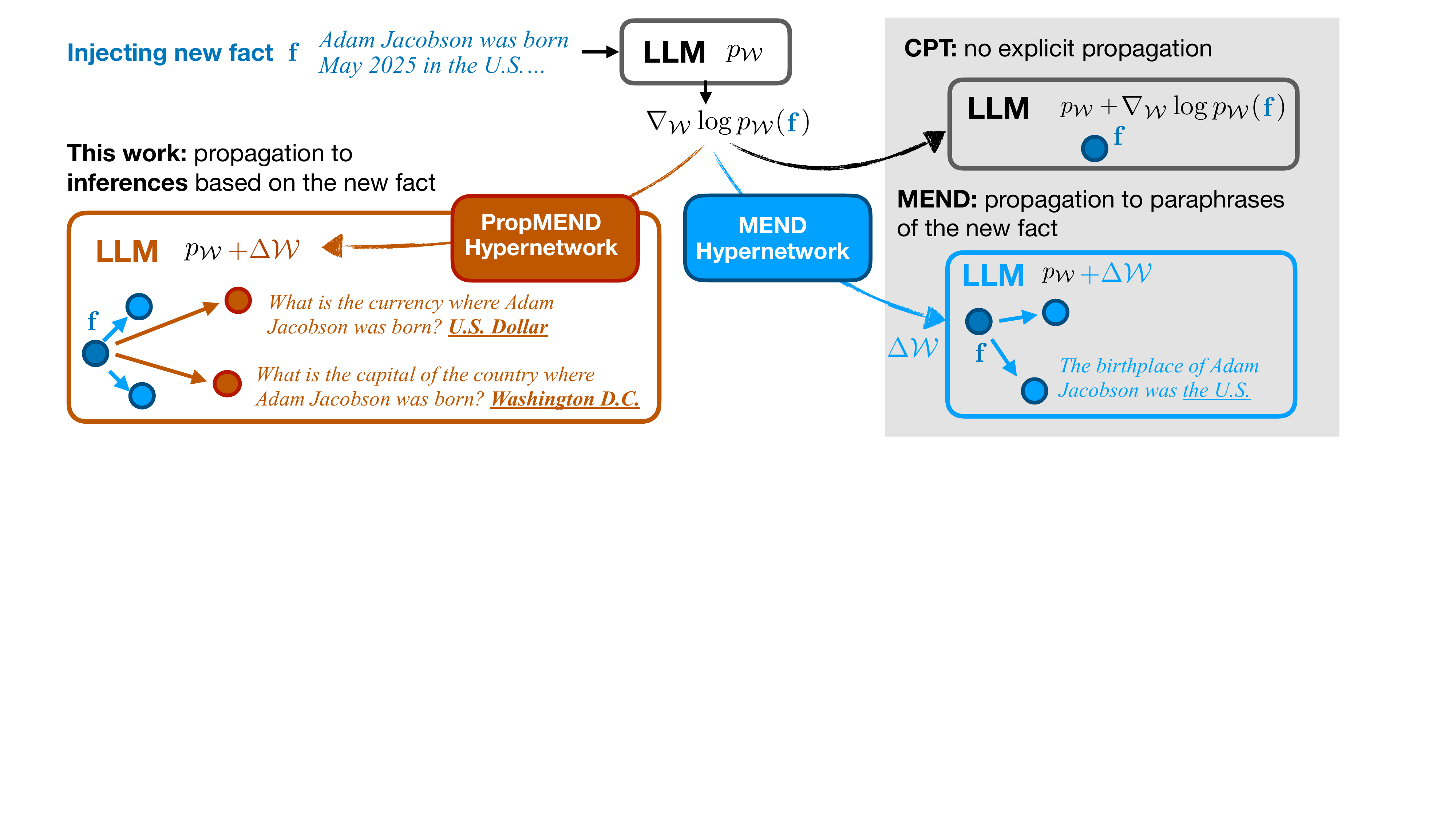}
  \caption{Our algorithm, \catchyname, enables the propagation of injected knowledge. Our hypernetwork is trained to modify the gradient from the next token prediction loss on the injected knowledge to allow answering of multi-hop questions that rely on the newly injected knowledge.}%
  \label{fig:main}
\end{figure}

In this work, we propose a new knowledge editing approach, named \catchyname, that achieves substantially improved results at knowledge propagation. Our method builds upon Model Editor Networks using Gradient Decomposition (MEND)~\cite{mend}, which introduces auxiliary hypernetworks to make efficient, local edits to LMs. We propose to train these hypernetworks with knowledge propagation as the core objective. Taking in a model's gradient from the language modeling objective on the injected fact as input, we train hypernetworks to modify that gradient to enable LMs to answer propagation questions involving that fact correctly when the output gradient is applied; see Figure~\ref{fig:main}. We further identify new settings of hyperparameters (e.g., layers in which model updates are applied) that improve the propagation performance significantly compared to MEND.

We first evaluate our approach on \realdata \cite{ripple_edit}, a knowledge propagation question answering dataset. Existing methods that excel in instances where the target answer appears verbatim in the injected facts, while achieving negligible improvement on non-verbatim questions. We show \catchyname outperforms all other approaches, showing almost $2\times$ accuracy (22.4\% compared to 12.7\% of the next best system) in non-verbatim cases. 

To better understand the extent of knowledge propagation, we design a new synthetic dataset \syndata. We focus on injecting facts related to well-known entities, allowing us to test propagation of information already known to LLMs. We design test sets to evaluate propagation relations and entities seen during hypernetwork training and those that are unseen. In this new dataset, we observe that our approach outperforms other approaches consistently, in both in-domain settings and on out-of-domain generalization. Our model performance is weakest in our hardest out-of-domain settings (17.7\% accuracy on propagation questions) compared to in-domain settings (64.0\%), indicating that further work on this benchmark can potentially develop even stronger methods to achieve generalization in knowledge propagation.  %

Our contributions are:
\begin{itemize}[noitemsep,leftmargin=10px]
    \item A new method for knowledge propagation, \catchyname{}, which meta-trains a hypernetwork explicitly for propagation.
    \item An analysis and evaluation on \realdata, showing that \catchyname{} achieves substantial improvement on questions whose answers are not verbatim in the injected fact. %
    \item A new dataset \syndata, which allows us to evaluate out-of-domain settings in knowledge propagation. Our model shows improvement over baselines in this challenging setting. %
    
\end{itemize}

The code and data is available at \texttt{\href{https://github.com/leo-liuzy/propmend}{https://github.com/leo-liuzy/propmend}}.
\section{Background}

\subsection{Task}
\label{sec:task}
We define a language model $\mathcal{M}$ with parameters $\mathcal{W}$ modeling a probability distribution $p_{\mathcal{W}}(x_i \mid \mathbf{x}_{<i})$ of current token $x_i$ given the previous tokens $\mathbf{x}_{<i}$. Such an LM is defined by its architecture and parameters, which are real-valued weight tensors $\mathcal{W} = \{W_{\ell,k}, \cdots \}$, where $\ell$ denotes the layer index and $k$ ranges over the number of weight types per layer (e.g., the MLP matrices and projection matrices for self-attention).

The task of knowledge editing is to inject a previously unknown fact or facts represented by $\mathbf{f}$ into the model. In this work, $\mathbf{f}$ consists of raw text (e.g., $\mathbf{f} =$\emph{``Keir Starmer was elected prime minister of the UK''}).
The weights are updated by $\Delta \mathcal{W} = \{\Delta W_{\ell,k}, \cdots \}$, yielding $\tilde{\mathcal{W}} = \{W_{\ell,k} + \Delta W_{\ell,k}, \cdots \}$ as the final weights which should reflect $\mathbf{f}$. Ideally, the model should be able to use this fact in various contexts (\emph{efficacy} of the edit) while maintaining \emph{locality} and not changing other unrelated facts.

We introduce a set of propagation questions associated with each injected set of facts: our data is of the form $\{(\mathbf{f}_i, \{(\mathbf{q}_{ij}, \mathbf{a}_{ij})\})\}$. For instance, given the $\mathbf{f}$ in the previous paragraph, propagation questions might be (\emph{Q: What year was the prime minister of the UK born? A: 1962}; \emph{What political party is the prime minister of the UK associated with? A: Labour Party}). 
These questions aims to evaluate that an updated language model should use its knowledge of the fact $\mathbf{f}$. Such questions have been explored in past work where they have been harvested from knowledge bases \cite{ripple_edit} or by prompting language models \cite{DCT}.

A natural approach is to compute an update to the weight $\Delta \mathcal{W}$ as the gradient of a language modeling loss or SFT loss computed on $\mathbf{f}$; for instance, $\Delta \mathcal{W} = \alpha \nabla p_{\mathcal{W}}(\mathbf{f})$, where $\alpha$ is the learning rate learned during meta-training. However, training a model on some text is typically insufficient to inject that knowledge in a way that leads to strong performance on the $(\mathbf{q}, \mathbf{a})$ pairs \cite{minjoon_fact, reversalCurse}.

\subsection{Hypernetwork-based Editing}
\label{sec:bgd:hypernet}

Our work builds on MEND~\cite{mend}, a hypernetwork-based method for knowledge editing. MEND computes an update $\Delta \mathcal{W}$ via a modification of the basic gradient.

\begin{figure}
  \centering
  \includegraphics[width=\textwidth,trim={25em 18em 5em 5em},clip]{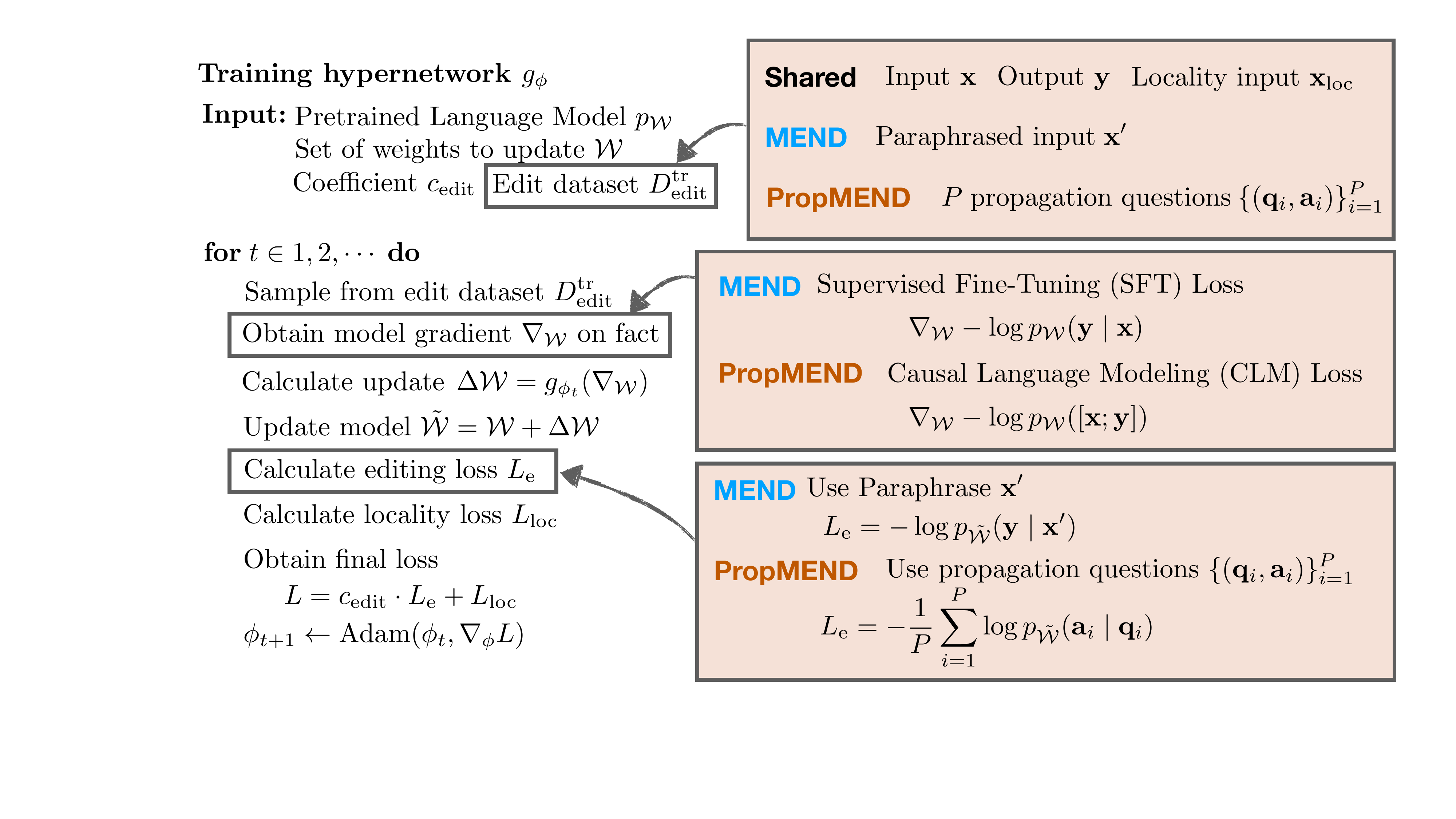}
  \caption{\catchyname. We learn a hypernetwork to take a gradient from causal language modeling of a new fact and transform it such that, when applied to the model, the model can answer propagations. The pseudocode skeleton follows MEND; differences between MEND and \catchyname are annotated.}
  \label{fig:algo:diff}
\end{figure}

The hypernetwork $g_{\phi}$ is parameterized by $\phi$ and meta-trained on an editing dataset $\dtrain = \{(\x, \y, \x', \xloc)_i\}$. As depicted in Figure~\ref{fig:algo:diff}, the training of the hypernetwork involves an inner-loop update which (1) computes the gradient of the injected fact; (2) modifies that gradient with the hypernetwork $g_\phi$; (3) applies the gradient to the base network $\mathcal{W}$ to form an updated network $\tilde{\mathcal{W}}$. In standard MEND, the gradient in (1) is computed over an input-output pair $(\mathbf{x}, \mathbf{y})$ (e.g., a QA pair) as  $\nabla_{\mathcal{W}} L^{I}(\x, \y) = \nabla_{\mathcal{W}} [- \log p_{\mathcal{W}}(\y \mid \x)]$. %

In the outer loop, the desiderata of generalization and locality is specified by a SFT loss (as editing loss $L_\text{e}$) with paraphrased input $\x'$ and Kullback–Leibler divergence (as locality loss $L_\text{loc}$) with a random input $\xloc$ from NaturalQuestion dataset~\cite{kwiatkowski2019natural}. An additional coefficient $c_\text{e}$ (typically $0.1$) is used to balance between the two desired properties.
\begin{equation}
    L^{O} = c_\text{e} L_{\text{e}}(\tilde{\mathcal{W}}) + L_{\text{loc}}(\mathcal{W}, \tilde{\mathcal{W}}) =  - c_\text{e} \log p_{\tilde{\mathcal{W}}}(\mathbf{y} \mid \mathbf{x}') + \text{KL}\left(p_\mathcal{W}(\cdot \mid \mathbf{x}_{\text{loc}}) \| p_{\tilde{\mathcal{W}}}(\cdot \mid \mathbf{x}_{\text{loc}} )\right)
\end{equation}

The full pseudocode for MEND can be found in Appendix~\ref{sec:details:mend}.
MEND makes a key observation that the gradient of $L^{I}$ with respect to weights $\mathcal{W}$ is a rank-1 matrix. This allows more efficient parameterization of the hypernetwork $g_\phi$ and efficient computation of the final weight update.

A major limitation of MEND is the structure of the inner- and outer-loop losses. As described in the paper, the inner loop injects a single QA pair $(\mathbf{x}, \mathbf{y})$, and the outer loop only encourages propagation to paraphrases of that QA pair. In the next section, we describe our method, which extends MEND and relaxes these assumptions.

\section{Method: \catchyname}
\label{sec:method}

\catchyname changes the training and loss of the MEND method, described below and visualized in Figure~\ref{fig:algo:diff}. There are two principal modifications (training data, learning objective) and other changes to the implementation to improve performance.

\paragraph{Meta-training} 
First, the loss in the outer loop is computed over the propagation questions:
\begin{equation}
    L_{\text{e}} = -\frac{1}{P} \sum_{i=1}^{P} \log p_{\tilde{\mathcal{W}}}(\mathbf{a}_i \mid \mathbf{q}_i)
    \label{eq:knowprop:outer:edit}
\end{equation}
Critically, this loss encourages the trained hypernetwork to make modifications that enable the final model to correctly answer propagation questions. This property does not hold for basic MEND; there, the objective in the outer loop is to predict simple paraphrases of the injected fact.

Second, we make the structure of the inner loop more flexible: we use the standard causal language model (CLM) loss to enable the model to inject any new knowledge expressible as text, rather than requiring it to be structured as QA pairs as in MEND:
\begin{equation}
    L^{I} = - \log p_{\mathcal{W}}([\mathbf{x}; \mathbf{y}]) = - \log p_{\mathcal{W}}(\mathbf{f})
    \label{eq:knowprop:inner}
\end{equation}
where $[\cdot \,; \cdot]$ means the concatenation of two strings. This objective resembles the inner loop loss used in past editing work \cite{reckoning}.

Together, these two losses reflect the chief objective of knowledge editing: taking raw knowledge expressed in text (which can be trained on with next token prediction loss) and adapting the learning of that knowledge to support answering propagation questions. 
This goal is more ambitious than that of MEND, which propagates QA pairs to paraphrases of those questions. MEND's injection may underperform on knowledge that is not expressed as QA pairs, and it may propagate less than a model explicitly trained to be able to answer propagation questions.

\paragraph{Hyperparameters} We re-investigate the hyperparameters and design choices of MEND, and we find that the choice of layers for parameter updating impacts the model's performance. MEND and other methods, such as MEMIT, selectively target certain layers within the LLM to modify. In MEND, the default configuration is to have the hypernetwork target the MLPs weights of the top 3 layers; however, we find editing lower layers is more effective for knowledge propagation. Applying the hypernetwork to all layers is expensive, since the hypernetwork operations are memory-intensive. Table~\ref{table:hyper:layers} in the appendix reports the layers modified with \catchyname.

\section{Evaluation on \realdata}
\label{sec:ripple}

\subsection{Experimental Settings}
\label{sec:ripple:exp}
\paragraph{Task}  In \realdata \cite{ripple_edit}, given an original \texttt{(subject, relation, object)} triplet $({s}, {r}, {o})$, an edit (e.g., ${o} \rightarrow  {o}^*$) is constructed to form a new triplet $\mathbf{e} = ({s}, {r}, {o}^*)$. The new triplet can be mapped into a natural language sentence with a template, which we denote as $\mathbf{f}$. Each edit can incur changes in other existing fact triplets.

\realdata captures propagation by identifying and preparing tests queries for 6 propagation types: 1. Logical Generalization (LG), a related fact that is created as a logical by-product of the relation $r$ (e.g., brother); 2. Compositionality I (CI), a multi-hop fact composed with another fact about the target object ${o}^*$; 3. Compositionality II (CII), a multi-hop fact that uses a different subject ${s}'$ but still holds for the new object ${o}^*$; 4. Subject Aliasing (SA), the same injected fact using paraphrased \texttt{subject-relation}; 5. Forgetfulness (FN), a neighbor triplet whose answer ${o}'$ does not change despite sharing the same relation ${r}$ as the edit (i.e., ${r}$ is a one-to-many relation); 6. Relation Specificity (RS), another fact about the subject ${s}$ that's not affected by the edits. See examples in Table~\ref{table:ripple:data}.

We evaluate on instances from \realdata with the following procedure. An LLM $\mathcal{M}$ receives an edited fact $\mathbf{e} = ({s}, {r}, {o}^*)$ to be injected into LLM, yielding an updated model $\mathcal{M}^{(\mathbf{e})}$. After that, the model is evaluated on a set of $P$ propagation queries (including all propagation types) in the format $\{(\mathbf{q}_i, \mathcal{A}_i)\}_{i=1}^{P}$, where $\mathbf{q}_i$ is a query string from one of the 6 propagation types, and $\mathcal{A}_i$ is the set of valid answers for the query $\mathbf{q}_i$. 

\paragraph{Data Setup} 
\realdata has three subsets, \texttt{Popular}, \texttt{Random}, and \texttt{Recent}. We do not distinguish these subsets for simplicity, and form the dataset splits out of the union of all of them. We randomly sample 500 examples for a validation set, 500 examples for a test set, and use the remaining 3,686 examples for training. We filter examples in the validation and test sets, such that each instance has at least 1 test query for efficacy and 1 test query for specificity. The training dataset here is used for meta-training our hypernetwork and not for learning of specific knowledge. See the statistics for a number of propagation questions in Table~\ref{table:stats:count}.

Following prior knowledge editing evaluations~\cite{sake}, we categorize six propagation types into two: (1) \textit{efficacy} queries (LG, CI, CII, SA), since these test the effectiveness of knowledge injection and propagation of a test fact. (2) \textit{specificity} queries (FN, RS), whose answer should not change after the edit. See illustration in Table~\ref{table:ripple:data:prop}.

Our analysis into the dataset revealed that answer to the propagated fact frequently appears verbatim in the edit fact (overall 31.9\% of propagation questions in test set; see breakdown per propagation type in Table~\ref{table:stats:spurious} in the Appendix). Models can trivially answer these questions correctly by learning to copy from edited facts. Therefore, we divide test queries into two sets: those that require \emph{non-verbatim propagation} and those that do not, and report performances on each set.

\paragraph{Evaluation Metrics} We greedily decode a maximum of 20 new tokens. We use two evaluation metrics, \textbf{Exact Match (EM)}, following the original paper, and \textbf{LLM-as-Judge (LLM-Score)}, a more robust metric that can handle lexical variations. \textbf{EM} checks if any gold answer $a \in \mathcal{A}_i$ is a substring of sequence $[\mathbf{q}_i; \hat{\mathbf{a}}_{i}]$ which concatenate the query string $\mathbf{q}_i$ with generated answer $\hat{\mathbf{a}}_{i}$.\footnote{In the original paper~\cite{ripple_edit}, the evaluation pipeline filters test queries based on edit success, performance on prerequisite test queries, making the set of evaluation queries different for different models. We do not filter to ensure each method is evaluated on the same test set.}   %
For \textbf{LLM-as-Judge (LLM-Score)}, an LLM (GPT-4o-mini) takes the query string $\mathbf{q}_i$, the generated answer $\hat{\mathbf{a}}_i$, and one answer from valid answers $a \in \mathcal{A}_i$, and gives a numerical score of whether the generated answer matches the valid answer. If the generated answer matches any of the valid answers, we count it as correct. See the LLM prompt in Appendix~\ref{prompt:llm-as-judge}.

\subsection{Comparison Systems} 
\label{subsec:compa}

All our model variants use the 16-layer transformer \texttt{Llama-3.2-1B-base} as its base architecture. Prompted with a question $q_i$, models will generate an answer followed by an end-of-sentence token. We conduct a light-weight supervised fine-tuning on the TriviaQA dataset~\cite{triviaqa} on this model to teach the model to answer in short answer format: $L_{\text{SFT}}(\mathcal{M}) = \E_{(\mathbf{x}, \mathbf{y}) \sim \text{TriviaQA}} \left[\log p_\mathcal{M}( \mathbf{y} \mid \mathbf{x})\right]$. We call the tune model \texttt{Llama-3.2-1B-base-QA}. %

\begin{itemize}[leftmargin=12px,noitemsep]
    \item{\textbf{Prepend}: This is not a knowledge editing method, simply prepending the new fact $\mathbf{f}$ to the test query $\mathbf{q}_i$ at inference time.  Past work has shown this method to be a competitive baseline \cite{ripple_edit, prop_by_distill,Onoe2023KnowledgeInject}.} 

    \item{\textbf{Continued Pretraining (CPT)} is frequently used to adapt an off-the-shelf LM to new domains or tasks~\cite{Gururangan2020DontSP}. We continue training the base model with the next token prediction loss (Equation~\ref{eq:knowprop:inner}) on the new fact $\mathbf{x}$. We report two variants, differing in which parameters are updated --- all parameters in the model (denoted CPT {\scriptsize(Full)}), or parameters associated with Layer-\texttt{[10-12]} (denoted CPT {\scriptsize(Mid-Upper)}).}
    
    \item{\textbf{MEMIT}~\cite{memit} requires precomputed covariance matrices from a reference corpus, typically on \texttt{wikitext-103} \cite{wiki103}. To reconcile potential train-test mismatch, we precompute the covariance matrix on the meta-training set of \catchyname, using both the injected facts and the propagation query-answer pairs. We denote MEMIT {\scriptsize(\texttt{wikitext-103})} to be MEMIT with covariance from \texttt{wikitext-103}, and MEMIT {\scriptsize(\texttt{\realdata})} to be from \realdata.  See more details in Appendix~\ref{sec:details}.}
    
    \item{\textbf{MEND}~\cite{mend}: We present two versions of MEND. MEND {\scriptsize(with standard config)} is trained on the zsRE question-answering dataset \cite{zsRE} with their original hyperparameters (editing top 3 MLP layers (i.e., Layer-\texttt{[13-15]})). Similar to our practice in MEMIT, we also change the meta-training set to be the meta-training set that \catchyname uses and targets at Mid-Upper Layers (denoted MEND {\scriptsize(Mid-Upper)}). This provides most controlled comparison setting with our method (same training dataset, same edit layers). We use \texttt{gpt-4o} to create a paraphrased input $\mathbf{x}'$ required for training. }
\end{itemize}

\subsection{Results}

\begin{table*}[t!]
\small
\centering
\addtolength{\tabcolsep}{-2.3pt}
\renewcommand{\arraystretch}{1.2}
\caption{\textbf{LLM-Score Results on \realdata dataset}. We report the total number of test queries in brackets. Our method \catchyname achieves improvement over the supervised fine-tuned model on verbatim questions whose answer is in the injected fact, and on non-verbatim questions whose answer is not in the injected fact. On the other hand, improvement of existing baselines mostly comes from improvement on the verbatim question. EM is reported in Table~\ref{tab:main:em} and performance by propagation types in Table~\ref{tab:main:all} in the appendix. \colorbox{gray!20}{Prepend} is not a parametric method. \mydagger means the system is outperformed by \catchyname on that metric according to a paired bootstrap test ($p=0.05$).} %
\label{tab:main:llm_acc}
\begin{tabular}{l cccc}
\toprule  
\multirow{3}{*}{LLM-Score $(\uparrow)$}  & \multicolumn{2}{c}{Efficacy }& \multicolumn{2}{c}{Specificity } \\
      &\begin{tabular}{@{}c@{}}Verbatim \\ {\tiny ($1373$)}\end{tabular}  & \begin{tabular}{@{}c@{}}Non-Verbatim \\ {\tiny ($1586$)}\end{tabular} &  \begin{tabular}{@{}c@{}}Verbatim \\ {\tiny ($165$)}\end{tabular}  & \begin{tabular}{@{}c@{}}Non-Verbatim \\ {\tiny ($2099$)}\end{tabular} \\
         \hline
\texttt{Llama-3.2-1B-base-QA}   & 11.6\mydagger & 9.2\mydagger & 13.2\mydagger & 27.7\mydagger \\
\rowcolor{gray!20} + Prepend & 36.7\mydagger & \textbf{22.4} & 18.8 & 28.7\mydagger \\
\hline
+ CPT {\scriptsize(Full)}  & \textbf{76.0}& 7.8\mydagger & 15.8\mydagger & 16.0\mydagger \\
+ CPT {\scriptsize(Mid-Upper)} & 41.8\mydagger & 9.7\mydagger & 20.7 & 26.3\mydagger \\
\begin{tabular}{@{}l@{}}+ MEMIT {\scriptsize(\texttt{wikitext-103})} \end{tabular}  & 17.0\mydagger & 12.7\mydagger & 17.7\mydagger & 24.5\mydagger \\
\begin{tabular}{@{}l@{}}+ MEMIT {\scriptsize(\realdata)} \end{tabular}  & 22.5\mydagger & 12.7\mydagger & 22.0 & 21.4\mydagger \\
+ MEND {\scriptsize(with standard config)} & 64.5\mydagger & 8.2\mydagger & \textbf{24.3} & 23.6\mydagger \\
+ MEND {\scriptsize(Mid-Upper)} & 63.5\mydagger  & 8.2\mydagger  & 21.6 & 21.6\mydagger \\

\hline
+ \catchyname\ {\scriptsize(Mid-Upper)} & 71.1\mydagger & 19.3\mydagger & 27.3 & 32.0\mydagger  \\
+ \catchyname\  & 75.7 & \textbf{22.4} & 24.1 & \textbf{35.4}  \\
\bottomrule
\end{tabular}
\end{table*}

Table~\ref{tab:main:llm_acc} presents the results on \realdata dataset. \catchyname performs strongly on both efficacy and specificity. Especially on non-verbatim questions, our system is the only one that shows substantial gain ($9.2 \rightarrow 22.4$), while the best other system achieves only $12.7$ (MEMIT). For existing methods, improvement in efficacy mostly comes from questions whose answer is verbatim in the edits ($11.6 \rightarrow 76.0$, CPT (full)), but offers negligible improvement on questions whose answers are not in the edits. On specificity questions, they show an increase on verbatim questions and decrease on non-verbatim questions. In contrast, Prepend improves on non-verbatim questions ($9.2 \rightarrow 22.4$) more substantially than other methods. %

\paragraph{Limitation of \realdata} While \realdata provides an initial testbed for our work, we find this dataset is not ideal for testing knowledge propagation. Many questions involve tail entities, where the base LM does not parametrically know the relevant information. For example, if LM does not know who are the siblings of Keir Starmer, it would not be able to answer the propagation question ``\emph{who is the sibling of the prime minister of the United Kingdom}" even if it could propagate the new fact ``\emph{Keir Starmer is the new PM of the UK}". In the following section, we present a new synthetic dataset that centers around entities and relationships that the model is familiar with. %

\section{Evaluation on \syndata}
\label{sec:synthetic}
We introduce a new dataset called \syndata, which will allow a focused evaluation of our model's knowledge propagation ability. We also design this dataset to evaluate out-of-domain performance, propagating along relations unseen during training, or with unseen entities. %

\begin{figure}
  \centering
  \includegraphics[width=\textwidth,trim={1em 56em 1em 2.5em},clip]{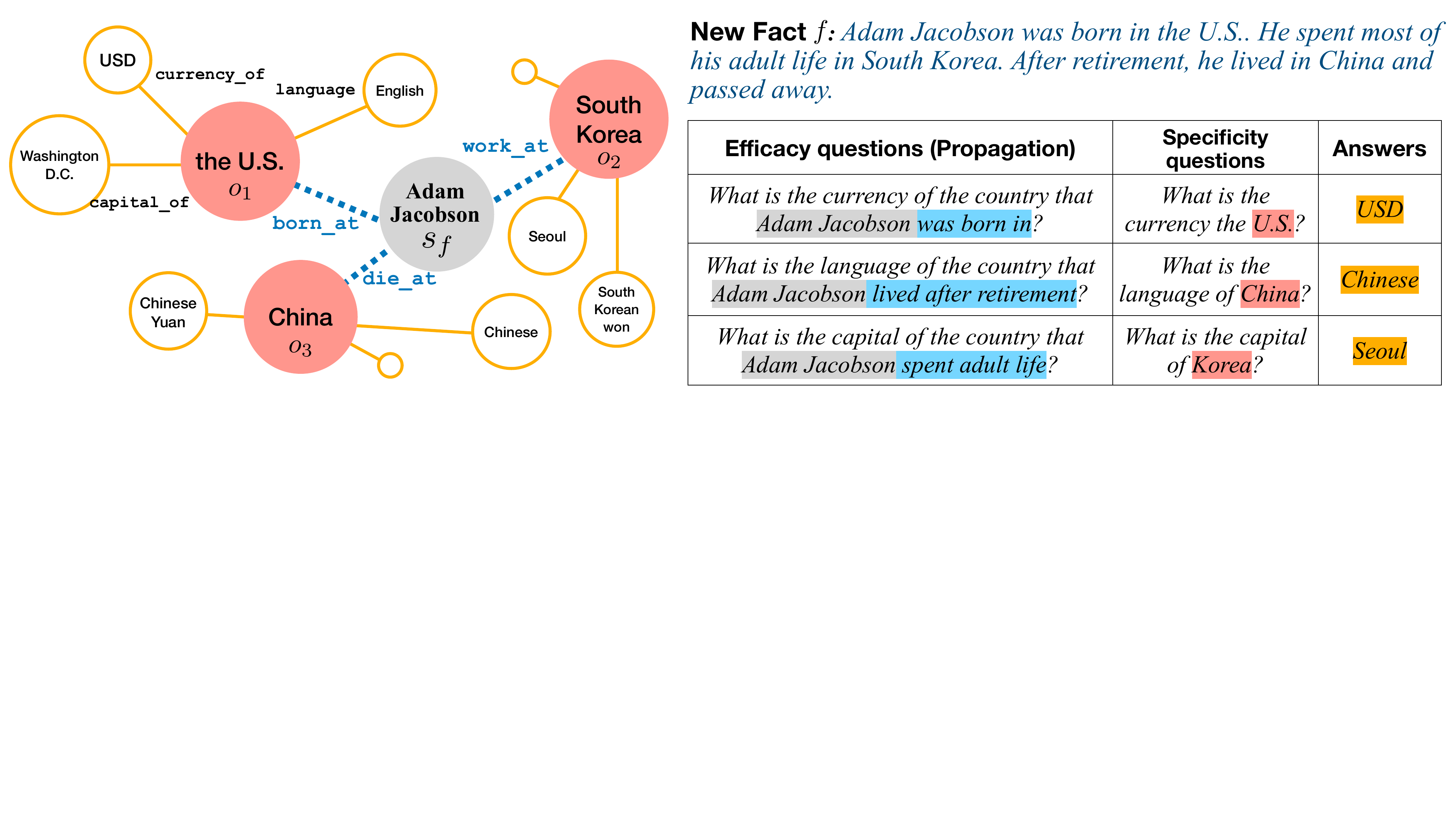}
  \caption{Illustration of our \syndata dataset, designed to evaluate knowledge propagation on well-known entities and relations. Each instance consists of (1) a fictional story ($\mathbf{f}$) relating a fake entity $s_f$ to three real-world entities ($o_1,o_2, o_3$); and (2) a set of $P$ propagation question-answer pairs $\{(\mathbf{q}_i, \mathbf{a}_i)\}_{i=1}^{P}$. Each $\mathbf{q}_{i}$ inquires about a knowledge base relation on one of the real-world entities $o_j$, but referring to it via its relation to the fake entity.}
  \label{fig:synstory}
\end{figure}

\paragraph{Data Instance} Figure~\ref{fig:synstory} illustrates an instance of \syndata. Each instance has a new fact $f$ centering around a fake entity $\mathbf{s}_f$ and involving three real-world entities $o_1, o_2, o_3$. It also has a set of propagation questions $\{(\mathbf{q}_i, \mathbf{a}_i)\}_{i=1}^{P}$ built from $P$ unique knowledge base relations (e.g., \texttt{capital\_of}) associated with one of the real-world entities ($o_1, o_2, o_3$). Instead of referring to the real world entity directly, the propagation question will refer to it using its relation to the fake entity $\mathbf{s}_f$ (e.g., \emph{the country where Adam Jacobson was born}). Therefore, the LM must be able to combine its prior knowledge about real-world entities and the injected fake entity $s_f$ to answer the question correctly.

\paragraph{Dataset Generation}
We manually select seven high-level categories for real-world entities: {person}, {event}, {language}, {creative work}, {organization}, {species}, and {country}. We manually design two fact templates per entity type, where one story template assumes the fake entity to be a person and the other a company. Figure~\ref{fig:synstory} shows an example where the type of the fake entity is person and the type of the real-world entity is country.

For each entity type, we prompt an LLM to generate (1) a list of entities belonging to the entity type and (2) relations relevant to the entity type. To effectively test propagation, we aim to restrict the entities and relations to those that are largely ``known'' by LLMs. Therefore, we filter datasets to obtain a smaller set of real-world entities (a total of 189 unique entities) and relations (a total of 38 unique relations).

From this set, we randomly sample three real-world entities of the same type and use fact template to generate fact to be injected. We can now form efficacy questions, querying relations on the real-world entities in the fact. The dataset generation process is further described in Appendix~\ref{sec:syndata:creation}.

\paragraph{Final Dataset}
We generate 5K instances of \syndata and randomly split these into 4K for training the hypernetwork, 500 for validation, and 500 for testing. To evaluate out-of-domain (OOD) generalization, we generate three additional test sets. We generate 350 instances where their real-world entities ($o_i$) do not appear in the training dataset (but knowledge base relations occur in the training dataset), calling this set OOD (Entity). Analogously, we generate an OOD (relation) dataset. Lastly, we generate an OOD (Both) dataset, consisting of 350 instances where neither real-world entities nor the knowledge base relations appear in the training dataset. %

\subsection{Experiment Setup}

\paragraph{Model} We use \texttt{Qwen-2.5-1.5B-base} instead of \texttt{Llama-3.2-1B-base} used in prior section, as we found the former showed much stronger performance in the Prepend setting. Similar to the previous section, we perform SFT on the TriviaQA dataset (see Section \ref{subsec:compa}) with \texttt{Qwen-2.5-1.5B-base} \cite{qwen25} to teach the question-answering format. We further train it with 500 QA pairs involving real-world entities and relations in \syndata to make the propagation easier by reinforcing the model's knowledge of the propagation relations. We call this model \texttt{Qwen-2.5-1.5B-base-QA}, and this model is used for all comparison methods in this section.

\paragraph{Metric} We use LLM-as-a-Judge (with GPT-4o-mini) to evaluate the correctness of the predicted answer against the reference answer, as in the prior section. For efficacy measure, we use model's performance on multi-hop questions, e.g., \textit{``Q: What is the currency of [the country that Adam Jacobson was born]? A: United States''}. To measure specificity, we evaluate whether the model retains its ability to answer simplified versions of our questions that do not require any updated knowledge, e.g., \textit{``What is the currency of the United States?''}. We refer to these as \textbf{single-hop questions}. See examples in Figure~\ref{fig:synstory}. Ideally, updates to the model should not degrade its ability to answer these questions.

\paragraph{Comparison Methods}
We use the same set of comparison methods described in Section~\ref{subsec:compa}. Since \texttt{Qwen-2.5-1.5B-base-QA} is a 28-layer transformer, we choose to edit Layer-\texttt{[18-22]} for \catchyname\ {\scriptsize(Mid-Upper)} and Layer-\texttt{[14-27(top)]} for \catchyname. For fair comparison, we modify MEMIT and MEND. As they require the fact $\mathbf{f}$ to be in an input-output format $(\mathbf{x}, \mathbf{y})$, we map $\mathbf{f}$ into three atomic facts (e.g., \textit{(Adam Jacobson was born in, the U.S.)}); and conduct multi-edit to inject those facts. See examples in Table~\ref{tab:syndata:example} and details in Appendix~\ref{sec:syndata:memit+mend}.

\input{tables/qwen1b_full_result_synstory}

\subsection{Results: Effectiveness of Propagation}
We report the results on \syndata in Table~\ref{tab:syndata:main:qwen}. \catchyname substantially outperforms other parametric methods consistently for various settings. On the in-domain test set, \catchyname even outperforms Prepend by $0.9\%$, showing that parametric propagation can be as effective as in-context augmentation. %

We observe \catchyname's performance degrades in out-of-domain settings when either entities or relations are unobserved during training. However, \catchyname still outperforms other methods substantially. For example, on OOD (Entity), the best-performing baseline MEMIT {\scriptsize(\texttt{wikitext-103})} achieves 18.6\% lower performance than \catchyname. We observe that \catchyname's performance improvement in OOD (Entity) tends to be higher than OOD (Relation). On OOD (Both), where \catchyname does not observe any entity or relation in the test, \catchyname is able to offer better propagation than others, but the gap is smaller.

\paragraph{Efficiency Evaluation} We report the efficiency of various editing methods, measured by their max memory usage and total runtime in Table~\ref{tab:efficiency:qwen}. ``Base Model'' does not involve any editing and only incurs inference costs. Different editing methods show different trade-offs between memory usage and runtime, and CPT {\scriptsize(Full)} is the least efficient in both dimensions. \catchyname is similarly efficient to MEND when editing the same number of layers, and gets less efficient when editing more layers. The number of layers being edited is the dominant factor in memory and runtime and outweighs the overhead due to the hypernetwork.

\input{tables/qwen1b_efficiency_synstory}

\paragraph{Ablation of \catchyname Design Choices}

Table~\ref{tab:syndata:main:qwen} presents ablations of the \catchyname design choices. First, we investigate having paraphrased inputs in the outer loop of \catchyname, similar to MEND, instead of propagation questions in the outer loop. This change is the most impactful one; without it, we see substantial performance degradation, suggesting that the hypernetwork training needs to be aligned with its intended test scenario. %
Second, we investigate changing the loss in the inner loop. In \catchyname, we apply the causal language modeling on all tokens of the fact $\mathbf{f}$. To change to SFT, we map  the fact $\mathbf{f}$ into three atomic facts taking an input-output format $(\mathbf{x}, \mathbf{y})$ (e.g., \textit{(Adam Jacobson was born in, the U.S.)}, see full example in Table~\ref{tab:syndata:example}); and the loss is calculated on the answer tokens $\mathbf{y}$ given the input $\mathbf{x}$. Training on all tokens as we do in \catchyname{} works substantially better in-domain, but in some OOD settings training on just answer tokens is competitive. Finally, we also find it is more effective to edit the Mid-Upper layers than the Upper layers of the transformer.

 \input{tables/qwen1b_scalup_synstory}

\input{tables/qwen1b_ablation_synstory}

\paragraph{Scaling up} We increase the hypernetwork size and the amount of meta-training data in Table~\ref{tab:syndata:scaleup:qwen} to investigate whether further scaling of the hypernetwork can lead to stronger performance. We find that increasing both can lead to substantial performance gains. %
However, although in-domain performance is close to perfect after scaling up both factors, increasing OOD performance remains a challenge.

\paragraph{Results with Other Base Models} We report experimental results with \texttt{Llama3.2-1B-base-QA} and \texttt{Llama3.2-3B-base-QA} in Table~\ref{tab:syndata:main:llama} and Table~\ref{tab:syndata:3b} in the appendix. We observe very similar experimental trends when editing \texttt{Qwen-2.5-1.5B-base-QA}, showing that the results from \catchyname hold for a different model family and size. We also conducted more extensive experiment with \texttt{Llama3.2-1B-base-QA}. See details in Appendix~\ref{sec:syndata:more-results}.

\section{Related work}

\paragraph{Knowledge Propagation} 
Recent work has studied the propagation of injected knowledge, finding that existing methods are largely lacking. A line of work~\cite{bidirection_edit,reversalCurse} studied reversal curse --- the model knows ``A is B'', but not ``B is A''. Other work~\cite{qin-etal-2024-new,nishi2025representation} analyzes unintended ripple effects of different editing methods. \citet{hase2024fundamental} surveys a wide range of open problems regarding revising the belief of the model. We discuss recent benchmarks for evaluating knowledge edits in Appendix~\ref{appendix:benchmark}.

\paragraph{Continual Learning} Knowledge editing can be viewed as continual learning, injecting new knowledge gradually. Continual learning has been studied in domain adaptation scenarios \cite{Gururangan2020DontSP, ke2023continual}. A line of work studies catastrophic forgetting during continual learning \cite{REMIX, franke2024preserving, jin2024demystifying, jin2024what}. They evaluate the performance on downstream tasks, rather than changes in parametric knowledge. 

Continued pretraining (CPT) on documents to be injected serves as a strong baseline in these scenarios. A line of work~\cite{prop_by_distill,DCT} proposes to improve knowledge propagation in CPT by modifying data scenarios or learning objectives. \citet{cake} uses circuit analysis to arrive at the template for data augmentation. \citet{jiang-etal-2024-instruction} finds instruction-tuning LMs on question-answering pairs prior to CPT is beneficial for knowledge injection. \citet{synCPT} proposes to synthesize large-scale data from the document to be injected and perform CPT on those documents, showing improved propagation. Unlike this line of work, \catchyname does not synthesize additional data at test time. 

\section{Conclusion}
\label{sec:conclusion}
In this work, we introduce \catchyname, a method that modifies slightly addresses the critical challenge of propagating edit to related fact in current knowledge editing techniques. 
We show the effectiveness of our method on \realdata, a widely-adopted dataset measuring propagation. We present a controlled dataset centering around well-known entities and and relations to further demonstrate the effectiveness when propagated knowledge is known by the model; we also show that our method maintains strong performance on out-of-domain test sets.

\paragraph{Limitations} Our study focuses on single-edit scenarios, and it is unknown how our method \catchyname would scale to multi-edit and multi-turn edit scenarios \cite{alphaedit, malmen_edit, RLEdit, zhang-etal-2024-dafnet, ma2025perturbationrestrained, hartvigsen2023aging, gu-etal-2024-model}. However, the hypernetwork could be optimized for multi-edit scenarios by incorporating multiple gradient updates in the inner loop. Our second limitation is parameter efficiency: our hypernetwork is as large as the edited language model. The limitation is inherited from MEND, but we believe it can be minimized further with future research. Finally, our work's evaluation is restricted to short-form answers, but evaluating on propagation for long-form answers would be valuable. In our preliminary study, we found if such answer is expected, \catchyname tend to degrade model's generation. 

\section*{Acknowledgments}
We thank Nicholas Tomlin, Fangcong Yin, Xi Ye, Hung-Ting Chen, Fangyuan Xu, and other members of UT NLP and NYU ML$^{2}$ for helpful feedback for earlier draft of this work.
This work was supported by the National Science Foundation under Cooperative Agreement 2421782 and the Simons Foundation grant MPS-AI-00010515 awarded to the NSF-Simons AI Institute for Cosmic Origins — CosmicAI, https://www.cosmicai.org/, a gift from Apple, a grant from Open Philanthropy, NSF CAREER Award IIS-2145280, and by the NSF AI Institute for Foundations of Machine Learning (IFML). This research has been supported by computing support on the Vista GPU Cluster through the Center for Generative AI (CGAI) and the Texas Advanced Computing Center (TACC) at the University of Texas at Austin. This work was done in part while the first and last author were visiting the Simons Institute for the Theory of Computing.

\bibliographystyle{plainnat}
\bibliography{references}

\section*{Appendix}

\newpage
\input{999_appendix}

\end{document}

%% file: commands.tex
\newcommand{\name}{MEND}

\newcommand{\x}{\mathbf{x}}
\newcommand{\y}{\mathbf{y}}

\newcommand{\xloc}{\mathbf{x}_\text{loc}}

\newcommand{\losse}{L_\text{e}}

\newcommand{\lossl}{L_\text{loc}}

\newcommand{\dtrain}{D^{tr}_{edit}}

\newcommand*{\fakemonochar}[1]{\makebox[.6em][c]{#1}}
\makeatletter
\newcommand*{\fakemonotext}[1]{%
  \@fakemonotext#1\\%
}
\newcommand*{\@fakemonotext}[1]{%
  \ifx #1\\\else\fakemonochar{#1}\expandafter\@fakemonotext\fi
}

\newif\ifcomments
\commentstrue
\ifcomments
\usepackage[normalem]{ulem}
\definecolor{CMpurple}{rgb}{0.6,0.18,0.64}
\newcommand\cm[1]{\textcolor{CMpurple}{\textsf{\scriptsize[\textbf{CM\@:} #1]}}}
\newcommand\cmi[1]{\textcolor{CMpurple}{#1}}
\newcommand\cmm[1]{\marginpar{\raggedright\tiny\textcolor{CMpurple}{\textsf{{\bfseries CM\@:} #1}}}}
\newcommand\cms{\bgroup\markoverwith{\textcolor{CMpurple}{\rule[.4ex]{2pt}{0.8pt}}}\ULon}
\else
\newcommand\cm[1]{}
\newcommand\cmi[1]{\ignorespaces}
\newcommand\cmm[1]{}
\newcommand\cms[1]{#1}
\fi

%% file: tables/qwen1b_full_result_synstory.tex
\begin{table}[t!]
\small
\centering
\addtolength{\tabcolsep}{-2.3pt}
\renewcommand{\arraystretch}{1.2}
\caption{Results on \syndata with \texttt{Qwen-2.5-1.5B-base-QA}. We report the model's LLM-Score on the dataset for efficacy, and the model's performance on a collection of single-hop questions for specificity. OOD (Entity) means using ID relation with OOD entity; OOD (Relation) means using ID entity with OOD relation. \colorbox{gray!20}{Prepend} is not a parametric method. \mydagger means the system is outperformed by \catchyname according to a paired bootstrap test ($p=0.05$).}
\label{tab:syndata:main:qwen}
\begin{tabular}{l|cc|cc|cc|cc}
\toprule
\multirow{2}{*}{LLM-Score $(\uparrow)$}  & \multicolumn{2}{c}{\begin{tabular}{@{}c@{}} In-Domain \\ ($2284$) \end{tabular}} & \multicolumn{2}{c}{\begin{tabular}{@{}c@{}} OOD (Entity) \\ ($1368$) \end{tabular}} & \multicolumn{2}{c}{\begin{tabular}{@{}c@{}} OOD (Relation) \\ ($421$) \end{tabular}}  & \multicolumn{2}{c}{\begin{tabular}{@{}c@{}} OOD (Both) \\ ($447$) \end{tabular}}\\
& Effi. & Spec. & Effi. & Spec. & Effi. & Spec. & Effi. & Spec. \\
\hline
                      
\texttt{Qwen-2.5-1.5B-base-QA} & 8.0\mydagger & 91.2\mydagger & 6.8\mydagger & 89.9
& 10.5\mydagger & \textbf{87.3} & 9.1\mydagger & \textbf{91.1} \\
\rowcolor{gray!20} + Prepend & 63.1 & 86.2\mydagger & \textbf{59.4} & 86.9 & \textbf{58.6} & 82.9 & \textbf{51.9} & 81.5\mydagger\\ \midrule
+ CPT (Full) & 12.0\mydagger & 88.2\mydagger & 9.6\mydagger &  86.8 & 12.0\mydagger & 82.7 & 11.2\mydagger & 82.0\mydagger \\
+ CPT  {\scriptsize(Mid-Upper)} & 8.4\mydagger & 91.2\mydagger & 6.9\mydagger & \textbf{90.3} & 10.6\mydagger & 87.2 & 10.4\mydagger & 90.4\\
+ MEMIT {\scriptsize(\texttt{wikitext-103})} & 16.0\mydagger & 91.3\mydagger & 16.1\mydagger & 90.1 & 13.9\mydagger & 87.2 & 9.6\mydagger & 90.3 \\
+ MEMIT {\scriptsize(\syndata)} & 11.6\mydagger & 91.2\mydagger & 12.6\mydagger & 90.0 & 10.3\mydagger & 86.6 & 10.1\mydagger & 89.7 \\

+ MEND {\scriptsize(with standard config)} & 12.3\mydagger & 87.1\mydagger & 9.9\mydagger & 88.2 & 11.1\mydagger & 83.5 & 10.9\mydagger & 86.2\\
+ MEND {\scriptsize(Mid-Upper)}  & 9.1\mydagger & 58.3\mydagger & 8.9\mydagger & 56.6\mydagger & 4.8\mydagger & 61.4\mydagger & 5.2\mydagger & 69.4\mydagger\\
\midrule
+ \catchyname\ {\scriptsize(Mid-Upper)} & 56.7\mydagger & 89.5\mydagger & 30.6\mydagger & 83.0 & 28.4\mydagger & 85.7 & 14.0\mydagger & 87.9 \\
+ \catchyname   & \textbf{64.0} & \textbf{93.6} & 34.7 & 83.0 & 33.3 & 84.8 & 17.7 &  85.8\\ 
\bottomrule
\end{tabular}
\end{table}

%% file: tables/qwen1b_efficiency_synstory.tex
\begin{table}[t!]
\small
\centering
\addtolength{\tabcolsep}{-2.3pt}
\renewcommand{\arraystretch}{1.2}
\caption{Efficiency Evaluation with \texttt{Qwen-2.5-1.5B-base-QA} model on 50 examples. All experiments are run on an NVIDIA GH200 120GB, in a server with a CPU of ARM Neoverse-V2. $^{\text{*}}$: we ran 4 gradient update on the injected fact $\mathbf{f}$, beyond which the drop in loss is marginal (see full hyperparameters in Table~\ref{table:hyper:cpt}).}
\label{tab:efficiency:qwen}
\begin{tabular}{l|c|c}

\toprule
           & Max Memory Usage (MiB $\downarrow$) & Total Runtime (Second $\downarrow$) \\
\hline
Base Model & 6763 & 61 \\
\rowcolor{gray!20} + Prepend  & + 20  & - 4  \\
\hline
+ CPT {\scriptsize (Full)$^{\text{*}}$} & + 25160 & + 1442 \\
+ MEMIT {\scriptsize(\texttt{wikitext-103})}  & + 4966 & + 1059 \\
+ MEND {\scriptsize(Mid-Upper)}  & + 8747 & + 111 \\
\hline
+ \catchyname\ {\scriptsize(Mid-Upper)} &   + 8741  & + 84 \\
+ \catchyname\  &   + 10217  & + 102 \\
\bottomrule
\end{tabular}
\vspace{-1em}
\end{table}

%% file: tables/qwen1b_scalup_synstory.tex
\begin{table}[t!]
\small
\centering
\addtolength{\tabcolsep}{-2.0pt}
\renewcommand{\arraystretch}{1.2}
\caption{Scaled-up experiment of \catchyname on \syndata with \texttt{Qwen-2.5-1.5B-base-QA}. We experiment with more in-domain meta-training instances, and different sizes of hypernetwork by having dedicated hypernetworks per target weight in \texttt{Qwen-2.5-1.5B-base-QA}. We observed that having larger training data and hypernetwork tends to improve performances on Out-of-Domain instances, but it remains challenging.}
\label{tab:syndata:scaleup:qwen}
\begin{tabular}{l|cc|cc|cc|cc|cc}
\toprule
  \multirow{2}{*}{LLM-Score $(\uparrow)$} & \multirow{2}{*}{\begin{tabular}{@{}c@{}} Hypernet \\ size (\# Param.) \end{tabular}} & \multirow{2}{*}{\begin{tabular}{@{}c@{}} \# train \\ instances \end{tabular}}  & \multicolumn{2}{c}{\begin{tabular}{@{}c@{}} In-Domain \\ ($2284$) \end{tabular}}  & \multicolumn{2}{c}{\begin{tabular}{@{}c@{}} OOD (Entity) \\ ($1368$) \end{tabular}} & \multicolumn{2}{c}{\begin{tabular}{@{}c@{}} OOD (Relation) \\ ($421$) \end{tabular}} & \multicolumn{2}{c}{\begin{tabular}{@{}c@{}} OOD (Both) \\ ($447$) \end{tabular}} \\

&  &  & Effi. & Spec. & Effi. & Spec. & Effi. & Spec. & Effi. & Spec. \\
\midrule
\multirow{2}{*}{\catchyname}  & \multirow{1}{*}{163M} & 4K & 64.0 & 93.6 & 34.7 & 83.0 & 33.3 & 84.8 & 17.7 &  \textbf{85.8}\\ 
& \multirow{1}{*}{3.4B} & 30K & \textbf{98.5} & \textbf{96.0}  & \textbf{42.2}  & \textbf{88.6}  & \textbf{42.9}  & \textbf{87.4}  & \textbf{17.8}  & 84.0 \\
\bottomrule
\end{tabular}
\vspace{-1em}
\end{table}

%% file: tables/qwen1b_ablation_synstory.tex
\begin{table}[t!]
\small
\centering
\renewcommand{\arraystretch}{1.2}
\caption{Ablation studies of \catchyname on \syndata with \texttt{Qwen-2.5-1.5B-base-QA}. To reduce compute costs, we run \catchyname\ {\scriptsize (Mid-Upper)}, which targets Layer-\texttt{[18-22]} for editing. ``Upper layer'' is Layer-\texttt{[23-27(top)]}. \mydagger means the system is out-performed by \catchyname\ {\scriptsize (Mid-Upper)} accroding to a paired bootstrap test ($p=0.05$).}
\label{tab:syndata:ablation}
\begin{tabular}{l|cc|cc|cc|cc}
\toprule
  \multirow{2}{*}{LLM-Score $(\uparrow)$} & \multicolumn{2}{c}{\begin{tabular}{@{}c@{}} In-Domain \\ ($2284$) \end{tabular}}  & \multicolumn{2}{c}{\begin{tabular}{@{}c@{}} OOD (Entity) \\ ($1368$) \end{tabular}} & \multicolumn{2}{c}{\begin{tabular}{@{}c@{}} OOD (Relation) \\ ($421$) \end{tabular}} & \multicolumn{2}{c}{\begin{tabular}{@{}c@{}} OOD (Both) \\ ($447$) \end{tabular}} \\

& Effi. & Spec. & Effi. & Spec. & Effi. & Spec. & Effi. & Spec. \\
\midrule

\catchyname\ {\scriptsize (Mid-Upper)} & \textbf{56.7} & 89.5 & \textbf{30.6} & 83.0 & \textbf{28.4} & 85.7 & 14.0 & 87.9 \\ 

\hspace{0.3cm}propagations $\rightarrow$ paraphrases & 10.6\mydagger & 89.9 & 9.3\mydagger & \textbf{90.4} & 12.6\mydagger & 84.6 & 10.2\mydagger & \textbf{88.3}  \\

\hspace{0.3cm}all tokens $\rightarrow$ answer tokens & 42.5\mydagger & \textbf{92.4}  & 30.0  & 89.0  & 22.7\mydagger  & \textbf{86.0}  & \textbf{14.7}  & 88.2 \\

\hspace{0.3cm}Mid-Upper $\rightarrow$ {Upper} layers  & 41.2\mydagger & 91.4 & 21.1\mydagger & 80.6\mydagger & 18.2\mydagger & 82.4\mydagger & 9.9\mydagger & 82.3\mydagger \\

\bottomrule
\end{tabular}
\vspace{-1em}
\end{table}

%% file: 999_appendix.tex
\appendix

\section{Prompt}

\begin{prompt}[title={\thetcbcounter{} LLM-as-Judge prompt}, label=prompt:llm-as-judge]

\char`[Instruction]\\
Please act as an impartial judge and evaluate the quality of the response provided by an AI assistant to the user question displayed below. For this evaluation, you should primarily consider the following criteria:\\
accuracy: \\
                Score 0: The answer is completely unrelated to the reference.\\
                Score 3: The answer has minor relevance but does not align with the reference.\\
                Score 5: The answer has moderate relevance but contains inaccuracies.\\
                Score 7: The answer aligns with the reference but has minor omissions.\\
                Score 10: The answer is completely accurate and aligns perfectly with the reference.\\
                Only respond with a numerical score.\\
\\
\char`[Question\char`]\\
\{question\}\\

\char`[The Start of Ground truth\char`]\\
\{reference\}\\
\char`[The End of Ground truth\char`]\\
\\
\char`[The Start of Assistant's Answer\char`]\\
\{prediction\}\\
\char`[The End of Assistant's Answer\char`]\\
\\
Return the numerical score wrapped in <score>..</score> tag

\end{prompt}

\section{Details on baseline methods}
\label{sec:details}
\subsection{Prepend}
\label{sec:details:prepend}
We follow the practice in \cite{ripple_edit} and format the prepended text to be ``\texttt{Imagine that } $\mathbf{f}$'', where $\mathbf{f}$ is the injected fact.

\subsection{MEMIT}
\label{sec:details:memit}
MEMIT \cite{memit} frames knowledge editing as an optimization problem to compute the updated weights. This method assumes three inputs: the verbalization of \texttt{subject-relation} $\mathbf{x}$, the string corresponding to \texttt{subject} ${s}$, and the string corresponding to \texttt{object} ${o}^*$. For the optimization to run effectively, the approach precomputes a covariance matrix (per target weight) from a reference corpus, typically, \texttt{wikitext-103} \cite{wiki103}. To reconcile potential train-test mismatch, we precompute the covariance matrix on the meta-training set of \catchyname, using both the injected facts, and the propagation query-answer pairs. See hyperparameters used in Appendix~\ref{sec:hyper}.

\subsection{MEND}
\label{sec:details:mend}

Our work follows the same hypernetwork structure as MEND \cite{mend}. We describe their design choices here, which are also adopted by our approach. Their algorithm is shown in Figure~\ref{fig:mend}.

\paragraph{Rank-1 matrix decomposition} Consider a specific weight matrix $W \in \mathcal{W}$. Let $\delta \in \mathbb{R}^m$ be the gradient of the loss with respect to the output of $W$; and $u \in \mathbb{R}^d$ be the input to the weight $W$. MEND observes that the gradient of the loss with respect to $W$, $\nabla_{\mathcal{W}} L^{I}$, is decomposable by the outer product between $\delta$ and $u$, namely $\delta {u}^{\top}$. The calculation can be extended to a batch instances via $\sum_{i=1}^{B} \delta^{i}  {u^{i}}^{\top}$, where superscipt $i$ denotes corresponding values for instance $i$. Due to this observation the hypernetwork $g_\phi$ parameterized by $\phi$ could operate on $\delta^{i}$ and $u^{i}$ as input without loss of information; correspondingly, it could output values ${\tilde{u}}$ and $\tilde{\delta}$ to compose the proposed update gradient through outer product $\tilde\nabla_{W} = \tilde{\delta} {\tilde{u}}^{\top}$. Finally, we compute $ W \gets W - \alpha \tilde\nabla_{W}$, where $\alpha$ is a learned weight-specific step size. This observation drastically reduces the computation cost of hypernetwork from $O(d\times m )$ to $O(d + m)$ and make training the hypernetwork feasible.

\paragraph{Parameter Sharing} 
When sharing is activated, gradients of the same shape (e.g., MLP down-projection in layer 10 and layer 12) will be modified by the same hypernetwork. To enable some layer-wise specialization, MEND applies a layer-specific scale and offset to the editor network hidden state and output, similar to FiLM layers \cite{perez2018film}. For the set of target weights $\mathcal{W}$, parameter sharing reduces computation costs of training the hypernetwork from $O(|\mathcal{W}| \cdot (d + m))$ to $O(c \cdot (d + m))$ for some constant $c$; in this study, since MLPs only have two distinct weight sizes (i.e., down-projection and up-projection), the constant $c = 2$. The recommended setting from MEND \cite{mend} is to do parameter sharing. We also follow the same setting.

\paragraph{MEND on \realdata} %
At test time, MEND uses Supervised Fine-Tuning loss to create the gradient input to the hypernetwork, with a verbalized prefix of subject-relation $(s, r, \cdot)$ as input and new object $o^*$ as output. To train the hypernet, one need paraphrase of $(s, r, \cdot)$. In the original setting, meta-training is conducted on the zsRE \cite{zsRE} dataset, which comes with paraphrasing. To make a more head-to-head comparison, we also train MEND on the meta-training set of \realdata, where we uses the same amount of data, all edit and propagation queries as the input, and we use \texttt{gpt-4o} to create missing paraphrases. 

\input{algs/mend}

\section{\realdata}
\label{sec:realdata:details}
The dataset is released under the MIT License, and is available at \url{https://github.com/edenbiran/RippleEdits/tree/main/data/benchmark}.

Table~\ref{table:ripple:data} shows examples of various propagation types. The example is adapted from \cite{ripple_edit}.
\input{tables/ripple_edit_example} In Table~\ref{table:stats:spurious}, we include a table showing what percentage of propagation questions per propagation type have one of their valid answers in the injected fact.

In Table~\ref{table:stats:count}, we include a table showing how many propagation questions are included per propagation type.

\input{tables/verbatim_test}

\begin{table}[]
    \centering
    \small
    \renewcommand{\arraystretch}{1.5}
    \caption{Verbatim rate on test examples. Percentage of \realdata propagation questions where one of the valid answers $a \in \mathcal{A}_i$ appeared in the edit fact in test examples.}
    \label{table:stats:count}
    \begin{tabular}{c|c|c|c}
    \toprule
     Total count & Train set & Validation set & Test set \\
    \hline
    \# Edit $(\mathbf{f}, \{(\mathbf{q}_i, \mathbf{a}_i)\})$ & 3686 & 500 & 500\\
    \hline 
    \# Logical Generalization questions & 2254 & 245 & 230  \\
    \# Compositionality I questions & 11045 & 1762 &  1679 \\
    \# Compositionality II questions & 1681 & 362 & 273  \\
    \# Subject Aliasing questions & 4898  & 715 & 777 \\
    \# Relation Specificity questions &  12223  & 2009 & 1982 \\
    \# Forgetfulness questions & 1881  & 304 &  282 \\
    \hline
    Overall & 33982  & 5397 & 5223 \\
    \bottomrule
    \end{tabular}
\end{table}

\begin{table}[]
\small
\centering
\addtolength{\tabcolsep}{-2.3pt}
\renewcommand{\arraystretch}{1.4}
\caption{An example instance of \syndata. As mentioned in Section~\ref{sec:syndata:memit+mend}, since some baselines require facts to be in input-output format, we also show an example for the processing.}
\label{tab:syndata:example}
\begin{tabular}{c|p{35em}}
\toprule        
\small
\multirow{2}{*}{$\mathbf{f}$}  &  \textit{{[Elizabeth Ruiz]}}{$s_f$} was born in \textbf{[Kenya]}$o_1$. She spent most of her adult life in \textbf{[Malaysia]}$o_2$. After retirement, she lived in \textbf{[Egypt]}$o_3$ and passed away.\\
\hline
\multirow{2}{*}{$\mathbf{q}_i, \mathbf{a}_i$}  & What is {the capital city of} the country that \textit{{[Elizabeth Ruiz]}}{$s_f$} spent most of her adult life in?, Kuala Lumpur\\
\hline
$\hat{\mathbf{q}}_i, \mathbf{a}_i$ & What is {the capital city of} \textbf{[Malaysia]}$o_2$?, Kuala Lumpur\\ 
\hline
\begin{tabular}{@{}c@{}}  3 Atomic facts
\\
$(\mathbf{x}, \mathbf{y})$ \end{tabular} & \begin{tabular}{@{}l@{}} ( \textit{{[Elizabeth Ruiz]}}{$s_f$} was born in, \textbf{[Kenya]}$o_1$ ) \\
( \textit{{[Elizabeth Ruiz]}}{$s_f$} spent most of her adult life at, \textbf{[Malaysia]}$o_2$ ) \\
( \textit{{[Elizabeth Ruiz]}}{$s_f$} died in, \textbf{[Egypt]}$o_3$ ) \end{tabular} \\
\bottomrule
\end{tabular}
\end{table}

\section{\syndata}\label{app:ref:data_constr}
In this section, we discuss implementation details regarding our controlled synthetic dataset \syndata. First, we discuss how we generate the components of our dataset (i.e., the well-known entities and relations) in Section~\ref{sec:syndata:creation}. Then, we describe how we conduct further filtering to a smaller set of entities and relations in Section~\ref{sec:syndata:small_filter}. We describe how we conduct additional preprocessing for baselines MEND and MEMIT in Section~\ref{sec:syndata:memit+mend}.

\subsection{Data Generation}
\label{sec:syndata:creation}
\paragraph{Generating the initial list of well-known entities and relations} We prompt ChatGPT to generate a list of head entities per entity type and manually filter out invalid entities. Then, starting from a list of general questions from ChatGPT, we manually iterate to obtain general relations per entity type. In generating the relation per entity type, we specifically aim for a general relation template that could be asked about any kind of entity within that type and could be answered with a short answer. Then, we programmatically generate all single-hop questions by instantiating each template with entity name. We prompt GPT-4.1 for answer or ``\emph{I don't know}''. After filtering for where answers are provided, we reprompt the model to shorten any answer that's longer than 30 characters. We treat the answer from GPT-4.1 as the gold answer; we observed this to be extremely reliable on instances that we manually inspected due to the well-known nature of the entities and relations. 

\paragraph{Generate facts and questions} Given a list of well-known entities and relations, we follow the following process in all cases to generate fact and its paired questions: (1) sample an entity type, where the probability of sampling an entity type determined by the number of entities of that type and whether that type has at least 1 relation; (2) randomly choose 3 entities from the list of entities of that type; (3) randomly choose which entity (among the 3 entities) to construct the efficacy and specificity question, for each relation of that entity type; (4) apply templates to arrive at facts and questions.

\subsection{Dataset Filtering}
\label{sec:syndata:small_filter}
We initially start with a set of 760 real-world entities and 48 relations. We filter this set to remove entities and relations not well-known to base LLMs. Specifically, we start with \texttt{Llama-3.2-1B-base-QA}) model. For each of 48 relations, we sample 10 real world entities and further train \texttt{Llama-3.2-1B-base-QA}) model with those 480 examples. 

With this model, we query all valid real-world entity, relation pairs. We use LLM-as-a-Judge to compare the predicted answer and GPT-4.1 answer, providing a score between 0 and 1. Then, we only keep pairs with LLM-as-a-Judge score higher than 0.4. For each entity type, all entities belonging to it have the same number of relations, the number of entities is at least 20, and the number of relation is at least 4. \textbf{In total, we end up with 189 entities and 38 relations (across entity types).} See the full list of entities in Table~\ref{table:syndata:small_entities}; see the list of relations in Table~\ref{table:syndata:small_relations} and the list of entities in Table~\ref{table:syndata:small_entities}.

\subsection{Baselines}
\label{sec:syndata:memit+mend}

\paragraph{Prepend}
We mildly modify the prompt from \cite{ripple_edit} to maintain grammaticality: for fake person as the subject, we use ``\texttt{Imagine that someone named } $\mathbf{f}$''; and for fake company as the subject, we use ``\texttt{Imagine that a company named } $\mathbf{f}$''.

\paragraph{Modifications for MEMIT and MEND}
MEMIT and MEND require the fact to be in an input-output format $(\mathbf{x}, \mathbf{y})$ and uses Supervised Fine-Tuning (SFT) loss $-\log p(\mathbf{y} \mid \mathbf{x})$, where output $\mathbf{y}$ is the real-world object $o_r$. For MEMIT, the input $\mathbf{x}$ is a verbalization for fake entity $s_f$ and the relation being tested $r$; and the name of the fake entity must be a substring of the verbalization. Although MEND does not require access to a substring of fake entity $s_f$, it requires a paraphrase of input $\mathbf{x}'$ for meta-training. Because story and question are template-generated, we also curate the templates to generate those components for each story template.

\input{tables/all_story_templates}

\input{tables/all_entities_synstory_small}

\input{tables/all_relations_synstory_small}

\section{\syndata Additional Results}
\label{sec:syndata:more-results}

\input{tables/llama1b_full_result_synstory}

In Table~\ref{tab:syndata:main:llama}, we include full test results with \texttt{Llama-3.2-1B-base-QA}. On the in-domain test set, \catchyname outperforms Prepend (the next best performing system) by $35.3\%$. We also observe performance degradation in out-of-domain settings. When either entities or relations are unobserved during training, \catchyname maintains a strong performance gap with other methods. For example, on OOD (Entity), the best-performing baseline CPT  {\scriptsize(Full)} achieves 18.2\% lower performance than \catchyname. Even on OOD (Both), where \catchyname does not observe any entity or relation in the test, \catchyname is able to offer slightly better propagation than others. Interestingly, we observe that OOD (Entity) performance tends to be higher than OOD (Relation), implying that entity and relation do not share the same level of difficulty for propagation.

In Table~\ref{tab:syndata:ablation:llama}, we show an ablation study with \catchyname, and observe similar finding as in Table~\ref{tab:syndata:ablation}. 

In Table~\ref{tab:efficiency:llama}, we present efficiency of various editing methods, measured by their max
memory usage and total runtime. The pattern is similar to what's observed in Table~\ref{tab:efficiency:qwen}.

In Table~\ref{tab:syndata:scaleup:llama}, we show experiment scaling up the size of hypernetwork and amount of meta-training data. This shows similar trends as observed in Table~\ref{tab:syndata:scaleup:qwen}.

\input{tables/llama1b_scalup_synstory}
\input{tables/llama1b_ablation_synstory}
\input{tables/llama1b_efficiency_synstory}

In Table~\ref{tab:syndata:3b}, results with \texttt{Llama-3.2-3B-base-QA} shows similar pattern in Table~\ref{tab:syndata:main:qwen}, and Table~\ref{tab:syndata:main:llama}.

\section{Hyperparameters}

\label{sec:hyper}

In Table~\ref{table:hyper:eos-sft}, we put the hyperparameters for supervised-finetuning conducted in our study to align model output format.

In Table~\ref{table:hyper:knowprop}, we put the hyperparameters for meta-training \catchyname and MEND. We mostly follows the default setting.

In Table~\ref{table:hyper:memit}, we put the hyperparameters for MEMIT. We mostly follows existing configurations in EasyEdit \cite{wang2023easyedit}.

In Table~\ref{table:hyper:cpt}, we put the hyperparameters for CPT baselines for both CPT {\scriptsize (Full)} and CPT {\scriptsize (Mid-Upper)}.

\begin{table}[]
	\centering
    \small
    \renewcommand{\arraystretch}{1.2}
	\setlength{\tabcolsep}{4pt}
    
 	\caption{Hyperparameters used for Supervised Fine-Tuning (SFT). The same set of parameters was used for \texttt{Llama-3.2-1B-base}, \texttt{Qwen-2.5-1.5B-base}, and \texttt{Llama-3.2-3B-base} (suffixed by \texttt{-QA}).}
    \label{table:hyper:eos-sft}

    \begin{subtable}{0.48\textwidth}
        \centering
        \caption{SFT on TriviaQA.rc.}
        \begin{tabular}{lc}
        \toprule
            Hyperparamter & Value  \\
            \hline
            Learning rate & 1e-5   \\
            Scheduler     & linear \\
            Epoch         & 2     \\
            Max seq. length & 256 \\
            Batch size & 128 \\
            Weight decay & 0.1 \\
            Max Gradient Norm & 1.0 \\
            WarmUp ratio & 0.03 \\
            Optimizer & AdamW \\
        \bottomrule
        \end{tabular}
    \end{subtable}
    \;\;\;
    \begin{subtable}{0.48\textwidth}
        \centering
        \caption{SFT on \syndata.}
        \begin{tabular}{lc}
        \toprule
            Hyperparamter & Value  \\
            \hline
            Learning rate & 2e-6   \\
            Scheduler     & linear \\
            Epoch         & 2     \\
            Max seq. length & 256 \\
            Batch size & 10 \\
            Weight decay & 0.1 \\
            Max Gradient Norm & 1.0 \\
            WarmUp ratio & 0.03 \\
            Optimizer & AdamW \\
        \bottomrule
        \end{tabular}
    \end{subtable}
    
\end{table}

\begin{table}[]
    \centering
    \small
    \renewcommand{\arraystretch}{1.2}
    \setlength{\tabcolsep}{4pt}
    
    \caption{Hyperparameters used for Continue Pretraining baselines, CPT {\scriptsize (Full)} and CPT {\scriptsize (Mid-Upper)}, when injecting one fact $\mathbf{f}$.}
    \label{table:hyper:cpt}
    \vspace{0.5em}
    \begin{tabular}{lc}
    \toprule
    Hyperparamter & Value  \\
    \hline
    Learning rate & 1e-5   \\
    Scheduler     & linear \\
    Epoch         & 4     \\
    Max seq. length & 1024 \\
    Batch size & 1 \\
    Weight decay & 0.1 \\
    Max Gradient Norm & 1.0 \\
    Optimizer & AdamW \\
    \bottomrule
    \end{tabular}
\end{table}

\begin{table}[]
    \centering
    \small
    \renewcommand{\arraystretch}{1.5}
    \setlength{\tabcolsep}{4pt}
    
    \caption{Hyperparameters used for \catchyname and MEND.}
    \label{table:hyper:knowprop}
    \vspace{0.5em}
    \begin{subtable}{0.4\textwidth}
        \centering
        \small
        \renewcommand{\arraystretch}{1.5}
        \setlength{\tabcolsep}{4pt}
        \caption{Hyperparameters for training \catchyname and MEND.}
        \label{table:hyper:training}
        \begin{tabular}{lc}
        \toprule
        Hyperparameter & Value  \\
        \hline
        $c_\text{edit}$ & 0.1 \\
        \begin{tabular}{@{}l@{}}learning rate to learn test-time learning rate $\alpha_\ell$\end{tabular} & 0.0001\\
        Learning rate for hypernetwork weight $\phi$ & 1.0e-06 \\
        Batch size (after gradient accumulation) & 10 \\
        Validation step & 100 \\
        Early stop patience (\# steps) & 2000 \\
        Maximum training step & 1000000 \\
        Optimizer & Adam\\
        \bottomrule
        \end{tabular}
    \end{subtable}
    \hfill
    \begin{subtable}{0.4\textwidth}
        \centering
        \small
        \renewcommand{\arraystretch}{1.5}
        \setlength{\tabcolsep}{4pt}
        \caption{Hyperparameters for hypernetwork (MLP) in \catchyname and MEND.}
        \label{table:hyper:arch}
        \begin{tabular}{lc}
        \toprule
        Hyperparameter & Value  \\
        \hline
        Activation & ReLU \\
        \# hidden & 1 \\
        \# hidden dim & 1920 \\
        \# parameter sharing & False \\
        \bottomrule
        \end{tabular}
    \end{subtable}
    \\
    \begin{subtable}{1\textwidth}
        \centering
        \small
        \renewcommand{\arraystretch}{1.5}
        \setlength{\tabcolsep}{4pt}
        \caption{Target MLP layers used for various comparison system}
        \label{table:hyper:layers}
        \begin{tabular}{lccc}
        \toprule
        Base Model & Total \# layers  & Comparison system & Layer indices (min: 0)  \\
        \hline
        \multirow{2}{*}{\texttt{Llama-3.2-1B-base}} & \multirow{2}{*}{16} & \catchyname &  4-15 \\
         & & \catchyname\ {\scriptsize (Mid-Upper)} / MEND {\scriptsize(Mid-Upper)} &  10-12 \\
         \hline
        \multirow{1}{*}{\texttt{Qwen2.5-1.5B-base}} & 28 & \catchyname &  13-27 \\
        \hline
        \multirow{1}{*}{\texttt{Llama-3.2-3B-base}} & 28 & \catchyname &  15-27 \\
        \bottomrule
        \end{tabular}
    \end{subtable}
\end{table}

\begin{table}[]
    \centering
    \small
    \renewcommand{\arraystretch}{1.5}
    \setlength{\tabcolsep}{4pt}
    
    \caption{Hyperparameters used for MEMIT.}
    \label{table:hyper:memit}
    \vspace{0.5em}
    \begin{subtable}{0.45\textwidth}
        \centering
        \small
        \renewcommand{\arraystretch}{1.5}
        \setlength{\tabcolsep}{4pt}
        \caption{For \texttt{Llama-3.2-1B-base}}
        \label{table:hyper:memit:llama}
        \begin{tabular}{lc}
        \toprule
        Hyperparameter & Value  \\
        \hline
        Target layer & [1, 2, 3, 4, 5] \\
        rewrite\_module\_tmp & ``layers.\{\}.mlp.down\_proj'' \\
        clamp\_norm\_factor & 0.75\\
        fact\_token & ``subject\_last''\\
        v\_num\_grad\_steps & 20 \\
        v\_lr & 5e-1 \\
        v\_loss\_layer & 15 \\
        v\_weight\_decay & 0.5 \\
        kl\_factor & 0.0625 \\
        mom2\_adjustment & true \\
        mom2\_update\_weight & 20000 \\
        mom2\_n\_samples & 100000\\
        \bottomrule
        \end{tabular}
    \end{subtable}
    \hfill
    \begin{subtable}{0.45\textwidth}
        \centering
        \small
        \renewcommand{\arraystretch}{1.5}
        \setlength{\tabcolsep}{4pt}
        \caption{For \texttt{Qwen-2.5-1.5B-base}}
        \label{table:hyper:memit:qwen}
        \begin{tabular}{lc}
        \toprule
        Hyperparameter & Value  \\
        \hline
        Target layer & [4, 5, 6, 7, 8] \\
        rewrite\_module\_tmp & ``layers.\{\}.mlp.down\_proj'' \\
        clamp\_norm\_factor & 4 \\
        fact\_token & "subject\_last"\\
        v\_num\_grad\_steps & 25 \\
        v\_lr & 5e-1 \\
        v\_loss\_layer & 27 \\
        v\_weight\_decay & 1e-3 \\
        kl\_factor & 0.0625 \\
        mom2\_adjustment & true \\
        mom2\_update\_weight & 15000 \\
        mom2\_n\_samples & 100000\\
        \bottomrule
        \end{tabular}
    \end{subtable}
\end{table}

\begin{table*}
\small
\centering
\addtolength{\tabcolsep}{-2.3pt}
\renewcommand{\arraystretch}{1.2}
\caption{\textbf{Exact Match (EM) Results on \realdata with \texttt{Llama-3.2-1B-base-QA}}. We report the total number of test queries in brackets. \colorbox{gray!20}{Prepend} is not a parametric method. The other metric (LLM-Score) is reported in Table~\ref{tab:main:llm_acc} in the main paper.} 
\label{tab:main:em}
\begin{tabular}{l cccc}
\toprule  
\multirow{3}{*}{EM $(\uparrow)$}  & \multicolumn{2}{c}{Efficacy }& \multicolumn{2}{c}{Specificity } \\
      &\begin{tabular}{@{}c@{}}Verbatim \\ {\tiny ($1373$)}\end{tabular}  & \begin{tabular}{@{}c@{}}Non-Verbatim \\ {\tiny ($1586$)}\end{tabular} &  \begin{tabular}{@{}c@{}}Verbatim \\ {\tiny ($165$)}\end{tabular}  & \begin{tabular}{@{}c@{}}Non-Verbatim \\ {\tiny ($2099$)}\end{tabular} \\
         \hline
\texttt{Llama-3.2-1B-base-QA}   & 17.0  & 4.0 & 90.9  & 23.2  \\
\rowcolor{gray!20} + Prepend & 36.0  & 12.4 & 94.5 & 21.6 \\
\hline
+ CPT {\scriptsize(Full)}  & 87.8 & 3.4 & \textbf{99.4} & 17.3  \\
+ CPT {\scriptsize(Mid-Upper)}&  48.7 & 4.0 & 93.3  &  24.1  \\
\begin{tabular}{@{}l@{}}+ MEMIT {\scriptsize(\texttt{wikitext-103})} \end{tabular}  &  21.1 & 5.6 &  93.3  &  24.1 \\
\begin{tabular}{@{}l@{}}+ MEMIT {\scriptsize(\realdata)} \end{tabular}  &  26.6 & 5.9 &  98.2  &  19.3  \\
+ MEND {\scriptsize(with standard config)} &  72.7 & 3.0 & 98.2 & 21.3  \\
+ MEND {\scriptsize(Mid-Upper)} & 69.7 & 3.1  &  97.0 &  17.8  \\
\hline
+ \catchyname\ {\scriptsize(Mid-Upper)} & 73.8  & 14.9 & 97.6 &  31.8 \\
+ \catchyname\  &  \textbf{78.7} & \textbf{17.3} &  95.2 &  \textbf{35.1} \\

\bottomrule

\end{tabular}
\end{table*}

\begin{table*}
\small
\centering
\addtolength{\tabcolsep}{-3pt}
\renewcommand{\arraystretch}{1.5}
\caption{\textbf{Results on \realdata with \texttt{Llama-3.2-1B-base-QA}}. Performances are reported in the format of Exact Match (EM) / LLM-Score. We notice the EM and LLM-Score strongly disagree with each other on Forgetfulness (FN); after spotchecking, we found EM is high because one of the valid answers $a \in \mathcal{A}_i$ is a substring of the propagation question $\mathbf{q}_i$. \colorbox{gray!20}{Prepend} is not a parametric method.}
\label{tab:main:all}
\begin{tabular}{l | cccccc}
\toprule
\multirow{3}{*}{EM / LLM-Score $(\uparrow)$}  &  \multicolumn{4}{c}{Efficacy} & \multicolumn{2}{c}{Specificity} \\
         \cmidrule(r){2-5} \cmidrule(r){6-7} 
        & \begin{tabular}{@{}c@{}}LG\\$(230)$\end{tabular} & \begin{tabular}{@{}c@{}}CI\\$(1679)$\end{tabular}  & \begin{tabular}{@{}c@{}}CII\\$(273)$\end{tabular} & \begin{tabular}{@{}c@{}}SA\\$(777)$\end{tabular} & \begin{tabular}{@{}c@{}}RS\\$(1982)$\end{tabular} & \begin{tabular}{@{}c@{}}FN\\$(282)$\end{tabular} \\
         \hline
\texttt{Llama-3.2-1B-base-QA} &  13.0$/$13.5  &  13.0$/$11.0  &  4.4$/$9.3  &  4.6$/$8.2  & 24.9$/$29.0 &  51.1$/$10.4 \\
\rowcolor{gray!20} + Prepend &  20.0$/$31.9 &  21.1$/$24.9  &  18.3$/$22.6  &  30.9$/$39.2  & 23.3$/$30.0 &  52.5$/$13.6 \\
\hline
+ CPT {\scriptsize(Full)}  &  16.1$/$11.4  &  12.7$/$10.4  &  93.8$/$89.3  &  97.0$/$93.0  & 19.9$/$17.8 &  47.5$/$3.3 \\
+ CPT {\scriptsize(Mid-Upper)}  &  13.9$/$15.8  &  13.3$/$12.0  &  32.6$/$32.2  &  50.1$/$51.7  & 26.4$/$28.0 &  48.6$/$10.9 \\
+ MEMIT {\scriptsize(\texttt{wikitext-103})}  &  14.3$/$13.8  &  14.5$/$14.6 &  7.3$/$11.6  &  10.6$/$16.2 & 24.1$/$26.3 &  49.6$/$7.9 \\
+ MEMIT {\scriptsize(\realdata)}  &  14.3$/$13.3  &  14.8$/$14.8  &  7.7$/$13.9  &  20.2$/$24.9  & 21.6$/$23.5 &  48.9$/$7.3 \\

+ MEND {\scriptsize(with standard config)} &  14.8$/$11.7  &  12.1$/$10.2  &  68.9$/$69.8  &  79.9$/$80.8  & 24.0$/$25.8 &  47.5$/$8.4 \\

+ MEND {\scriptsize(Mid-Upper)} &  13.5$/$13.8  &  12.4$/$10.8  &  59.0$/$64.1  &  77.9$/$79.2  & 20.1$/$23.6 &  47.5$/$8.1 \\
\hline
+ \catchyname\ {\scriptsize(Mid-Upper)} &  27.0$/$12.8  &  22.9$/$25.9  &  72.5$/$74.3  &  77.7$/$79.3  & 33.3$/$33.1 &  59.9$/$21.5 \\
+ \catchyname\ &  30.9$/$25.0  &  25.3$/$27.7  &  83.5$/$85.7  &  81.3$/$82.1  & 35.7$/$35.6 &  65.6$/$27.3 \\

\bottomrule
\end{tabular}
\label{table:mteb}
\end{table*}

\begin{table}[]
\small
\centering
\addtolength{\tabcolsep}{-2.3pt}
\renewcommand{\arraystretch}{1.2}
\caption{Results on \syndata with \texttt{Llama-3.2-3B-base-QA}. We use the model's LLM-Score on multi-hop questions for efficacy, and the model's performance on single-hop questions for specificity. OOD (Entity) means using ID relation with OOD entity; OOD (Relation) means using ID entity with OOD relation. \colorbox{gray!20}{Prepend} is not a parametric method. }
\label{tab:syndata:3b}
\begin{tabular}{l|cc|cc|cc|cc}
\toprule
\multirow{2}{*}{LLM-Score $(\uparrow)$}  & \multicolumn{2}{c}{\begin{tabular}{@{}c@{}} In-Domain \\ ($2284$) \end{tabular}} & \multicolumn{2}{c}{\begin{tabular}{@{}c@{}} OOD(Entity) \\ ($1368$) \end{tabular}} & \multicolumn{2}{c}{\begin{tabular}{@{}c@{}} OOD(Relation) \\ ($421$) \end{tabular}}  & \multicolumn{2}{c}{\begin{tabular}{@{}c@{}} OOD(Both) \\ ($447$) \end{tabular}}\\
& Effi. & Spec. & Effi. & Spec. & Effi. & Spec. & Effi. & Spec. \\
\hline
                      
\texttt{Llama-3.2-3B-base-QA} & 8.1 & 91.8 & 6.9 & 93.0 & 8.1 & 92.4 & 6.5 & 93.8 \\
\rowcolor{gray!20} + Prepend & 66.1 & 90.3 & 62.5 & 92.1 & 61.3 & 90.3 & 52.5 & 91.6\\ \midrule
+ CPT (Full) & 18.4 & 86.2 & 16.8 & 86.0 & 16.1 & 86.7 & 12.7 & 82.7 \\
\midrule
+ \catchyname  &  69.9  & 94.6 & 42.4 & 89.8 & 34.0 & 93.2 & 19.2 & 89.6 \\ 
\bottomrule
\end{tabular}
\end{table}

\section{Other propagation benchmarks} \label{appendix:benchmark}

Other benchmarks have attempted to capture knowledge propagation. DeepKnowledge \cite{deepknowledge} is a concurrent dataset testing propagation at various levels, but this dataset is not yet released at the time of development. MQuake and its improved version MQuake-Remastered \cite{zhong2023mquake, zhong2025mquakeremastered} aim at capturing propagation by testing whether the model is able to conduct multi-hop reasoning. In our preliminary study, we also considered a multi-hop question answering dataset for our study, but we found 100\% verbatim rate from instances in MQuake-Remastered. A similar issue exists in MuSiQue \cite{trivedi-etal-2022-musique} and other multi-hop question answering datasets \cite{yang-etal-2018-hotpotqa}. \citet{Onoe2023KnowledgeInject, Onoe2022EntityCB} study the task of learning a new entity through description (e.g., ``\textit{Dracula}''), and ask inference questions about the entity (e.g., ``Dracula makes you  \textit{fear}''). CodeUpdateArena \cite{codeupdatearena} tests whether the model could learn a function update in the docstring difference and apply the updated function in program synthesis. ECLeKTic \cite{ECLeKTic} focuses on cross-lingual knowledge transfer.

\section{Computational resources}\label{sec:compute-res}

We conducted experiments with \texttt{Llama-3.2-1B-base} primarily on a server with NVIDIA A40 48GB GPUs and an AMD EPYC 7413 24-Core Processor. For larger models, our experiments were conducted on a server with NVIDIA GH200 120GB and ARM Neoverse-V2. 

Though the runtime varies depending on the datasets, the meta-training of hyper networks typically takes around 10 hours, or as little as 4 hours for some experiments.

%% file: algs/mend.tex
\begin{figure}
\caption{MEND algorithm; reproduced from \cite{mend}}
\label{fig:mend}
\vspace{-1.8em}
\begin{minipage}[t]{0.43\textwidth}
\begin{algorithm}[H]
\small
\caption{MEND Training (\underline{Outer} Loop)}\label{alg:mend:training}
\begin{algorithmic}[1]
\State \textbf{Input:} Pre-trained $p_{\theta}$, weights to make editable $\mathcal{W} \subseteq \theta$, editor params $\phi$, edit dataset $\dtrain$, edit-locality tradeoff $c_\text{edit}$
\For{$t \in 1,2,...$}
\State Sample $\x,\y, \mathbf{x}', \xloc \sim \dtrain$
\State $\mathcal{\tilde W} \gets \textsc{Edit}(\theta, \mathcal{W},\phi_{t-1},\x, \y)$ 
\State $\losse \gets -\log p_{\mathcal{\tilde W}}(\y \mid \x')$
\State $\lossl \gets \text{KL}(p_{\mathcal{W}}(\cdot \mid \xloc) \| p_{\mathcal{\tilde W}} (\cdot\mid\xloc))$ 
\State $L^{O}(\phi_{t-1}) \gets c_\text{edit} \losse + \lossl$
\State $\phi_{t} \gets \text{Adam}\left(\phi_{t-1}, \nabla_\phi L(\phi_{t-1})\right)$
\EndFor
\end{algorithmic}\end{algorithm}\end{minipage} \hfill%
\begin{minipage}[t]{0.56\textwidth}
\begin{algorithm}[H]
\small
\caption{\name~Edit Procedure (\underline{Inner} Loop)}\label{alg:mend:edit-procedure}
\begin{algorithmic}[1]
\Procedure{Edit}{$\theta,\mathcal{W},\phi,\x,\y$}
    \State $\hat{p} \gets p_{\theta}(\y \mid \x)$, \textbf{caching} input $u_\ell$ to $W_\ell\in\mathcal{W}$
    \State $L^{I}(\x,\y)\gets -\log \hat{p}$ \hspace{-3em} \Comment{Compute neg log-likelihood}
    \For{$W_\ell \in \mathcal{W}$}
        \State $\delta_{\ell+1} \gets \nabla_{W_\ell u_\ell} L^{I}(\x,\y)$ \Comment{Grad w.r.t. output}
        \State $\tilde u_\ell,\tilde\delta_{\ell+1} \gets g_{\phi_\ell}(u_\ell,\delta_{\ell+1})$ \hspace{-2em} \Comment{Rank-1 udpate vec}
        \State $\tilde\nabla_{W_\ell} \gets \tilde\delta_{\ell+1}\tilde u_\ell^{\top}$ \hspace{-2em} \Comment{Compose the full update grad}
        \State $\tilde W_\ell \gets W_\ell - \alpha_\ell \tilde\nabla_{W_\ell}$ \Comment{Learned step size $\alpha_\ell$}
    \EndFor
    \State $\mathcal{\tilde{W}} \gets \{\tilde W_1,...,\tilde W_k\}$; \Return $\mathcal{\tilde{W}}$ 
\EndProcedure
\end{algorithmic}\end{algorithm}\end{minipage}
\end{figure}

%% file: tables/ripple_edit_example.tex
\begin{table*}[]
\centering
\small
\renewcommand{\arraystretch}{1.4}
\setlength{\tabcolsep}{4pt}
\caption{\realdata example across various propagation types. The example is adapted from \cite{ripple_edit}.}
\label{table:ripple:data}
\begin{subtable}{1\textwidth}
    \centering
    \caption{A snapshot of world knowledge at the time of edit.}
    \label{table:ripple:data:given}
    \begin{tabular}{ll}
    \toprule
        Entity & Knowledge Triplets  \\
        \hline
        \multirow{5}{*}{\begin{tabular}{@{}l@{}}Prince \\ \circled{4} (Prince, \texttt{alias}, Prince Roger Nelson)\end{tabular}}
        & \circled{1} (Prince, \texttt{sibling}, Tyka Nelson) \\
        & \circled{2} (Tyka Nelson, \texttt{profession}, Singer) \\
        & \circled{3} (Prince, \texttt{founder\_of}, Paisley Park Records) \\
        & \circled{5} (Mattie Shaw, \texttt{mother\_of}, Prince) \\
        \hline
        \multirow{2}{*}{\begin{tabular}{@{}l@{}}Nicholas Carminowe\\\end{tabular}}
        & \circled{6} (Nicholas Carminowe, \texttt{profession}, Members of Parliament) \\
        & \circled{7} (Nicholas Carminowe, \texttt{sibling}, John Carminowe) \\
    \bottomrule
    \end{tabular}
\end{subtable}
\\\vspace{1em}
\begin{subtable}{1\textwidth}
    \centering
    \caption{Edit that introduce changes among entities.}
    \label{table:ripple:data:edit}
    \begin{tabular}{c}
    \toprule
    New relation created  \\
    \hline
    \circled{8} (Prince, \texttt{sibling}, Nicholas Carminowe) \\
    \bottomrule
    \end{tabular}
\end{subtable}
\\\vspace{1em}
\begin{subtable}{1\textwidth}
    \centering
    \caption{Propagation that follows from the edit in Table~\ref{table:ripple:data:edit}. We highlight the use of injected fact \textbf{\circled{8}}, and the cases where certain knowledge is expected to be \textbf{[Not forgotten]}.}
    \label{table:ripple:data:prop}
    \begin{tabular}{lp{4.7cm}ll}
    \toprule
    Propagation type & Question & Answer (Explanation) \\
    \hline
    \begin{tabular}{@{}l@{}}Logical\\Genralization\end{tabular} & The siblings of Nicholas Carminowe are & \begin{tabular}{@{}l@{}}  Prince (\textbf{\circled{8}} + \texttt{sibling} is a symmetric relation)\\ John Carminowe (\circled{6})\end{tabular}  \\
    \hline
    \multirow{2}{*}{Compositionality I} & The professions of the siblings of Prince are & \begin{tabular}{@{}l@{}}  Members of Parliament (\textbf{\circled{8}} + \circled{5}) \\ Singer (\circled{1} + \circled{2})\end{tabular} \\
    \hline
    \multirow{2}{*}{Compositionality II} & The siblings of the founder of Paisley Park Records are & \begin{tabular}{@{}l@{}} Nicholas Carminowe (\circled{3} + \textbf{\circled{8}})\\ Tyka Nelson (\circled{3} + \circled{1})\end{tabular}\\
    \hline
    Subject Aliasing & The siblings of Prince Roger Nelson are & \begin{tabular}{@{}l@{}} Nicholas Carminowe (\circled{4} + \textbf{\circled{8}})\\ Tyka Nelson (\circled{4} + \circled{1})\end{tabular}\\
    \hline
    Forgetfulness & The siblings of Prince are & \begin{tabular}{@{}l@{}} Nicholas Carminowe (\textbf{\circled{8}}) \\ Tyka Nelson (\circled{1}) \textbf{[Not forgotten]}\end{tabular}\\
    \hline
    Relation Specificity & The mother of Prince is & \begin{tabular}{@{}l@{}} Mattie Shaw (\textbf{\circled{8}}) \textbf{[Not forgotten]}\end{tabular}\\
    \bottomrule
    \end{tabular}
\end{subtable}
\\

\end{table*}

%% file: tables/verbatim_test.tex
\begin{table}[]
    \centering
    \small
    \renewcommand{\arraystretch}{1.5}
    \caption{ Percentage of verbatim question in \realdata, where the one of the valid answers $a \in \mathcal{A}_i$ appeared in the edit fact in test examples.}
    \label{table:stats:spurious}
    \begin{tabular}{c| c | c| c}
    \toprule
    Propagation Query Type & Train set & Validation set & Test set \\
    \hline
    Percentage of verbatim question in Logical Generalization & 35.8\% & 51.8\% & 55.2\% \\
    Percentage of verbatim question in Compositionality I & 11.0 & 12.3\% & 11.7\% \\
    Percentage of verbatim question in Compositionality II & 100.0\% & 100.0\% & 100\% \\
    Percentage of verbatim question in Subject Aliasing & 100.0\% & 100.0\% & 100\% \\
    Percentage of verbatim question in Relation Specificity & 3.2\% & 3.5\% & 3.2\% \\
    Percentage of verbatim question in Forgetfulness & 87.4\% & 79.3\% & 81.9\% \\
    \hline
    Overall & 31.3\% & 32.1\% & 31.9\%   \\
    \bottomrule
    \end{tabular}
\end{table}

%% file: tables/all_story_templates.tex
\begin{table*}[]
\centering
\resulttablefontsize
\renewcommand{\arraystretch}{1.2}
\setlength{\tabcolsep}{4pt}
\caption{Story templates of all entity types.}
\label{table:syndata:story_template}
\begin{tabular}{llp{0.6\linewidth}}
\toprule
Real-world Entity Type   & Subject Type & Story Template       \\
\hline

\multirow{6}{*}{Country} & Person       &   \{subject\} was born in \{country\_1\}. \{Gender\_subj\} spent most of \{gender\_possessive\_adj\} adult life in \{country\_2\}. After retirement, \{gender\_subj\} lived in \{country\_3\} and passed away. 
     \\
\cmidrule(r){2-3}
                         & Company      &  \{subject\} was founded in \{country\_1\}. \{Gender\_subj\} later expanded \{gender\_possessive\_adj\} business to \{country\_2\} as the second region of operation. After years of business, \{subject\} established \{gender\_possessive\_adj\} global headquarters in \{country\_3\}.
 \\
\hline
\multirow{6}{*}{Person}  & Person       &  \{subject\} first wrote about \{person\_1\} in an 8th-grade book report. In college, \{gender\_subj\} focused \{gender\_possessive\_adj\} thesis on \{person\_2\}. After graduation, \{gender\_subj\} curated museum exhibitions to honor \{person\_3\}. \\
\cmidrule(r){2-3}
                         & Company      &  \{subject\} drew inspiration from \{person\_1\} when shaping \{gender\_possessive\_adj\} mission. Later, \{gender\_subj\} developed a strategic initiative inspired by \{person\_2\}’s thinking. Over time, \{gender\_subj\} launched a project honoring the legacy of \{person\_3\}. \\
\hline
\multirow{7}{*}{Event}   & Person       & \{subject\} developed a passion for history after learning about \{event\_1\} in grade school. In college, \{gender\_subj\} did research on \{event\_2\}. Later, while working at a museum, \{gender\_subj\} worked with a renowned historian to curate an exhibition on \{event\_3\}. \\
\cmidrule(r){2-3}
                         & Company      & \{subject\} drew early inspiration from \{event\_1\} to shape \{gender\_possessive\_adj\} culture. Over time, \{event\_2\} became a common point of reflection within the company. Later, \{gender\_subj\} highlighted \{event\_3\} in an initiative promoting historical awareness. \\
\hline
\multirow{8}{*}{Species}   & Person       & \{subject\} became fascinated with nature after learning about \{species\_1\}. During graduate school, \{gender\_subj\} researched on \{species\_2\}. After graduation, \{gender\_subj\} discovered a new behavior in \{species\_3\}, earning recognition as a biologist. \\
\cmidrule(r){2-3}
                         & Company      & \{subject\} developed an interest in wildlife while supporting a conservation project for \{species\_1\}. \{Gender\_subj\} later partnered with researchers to study \{species\_2\}. \{Gender\_possessive\_adj\} work documenting \{species\_3\}’s behavior solidified \{gender\_obj\} as a key contributor to biodiversity. \\
\hline
\multirow{6}{*}{Language}   & Person       & \{subject\} was born into a \{language\_1\}-speaking environment. In grade school, \{gender\_subj\} started to learn \{language\_2\}. In \{gender\_possessive\_adj\} college, \{gender\_subj\} took a major in \{language\_3\}. \\
\cmidrule(r){2-3}
                         & Company      & \{subject\} began by offering services in \{language\_1\}. \{Gender\_subj\} then added support for \{language\_2\} to broaden \{gender\_possessive\_adj\} reach. Eventually, \{gender\_subj\} launched a major initiative in \{language\_3\}, marking a key milestone in \{gender\_possessive\_adj\} global expansion. \\
\hline
\multirow{6}{*}{Organization}   & Person       & \{subject\} began \{gender\_possessive\_adj\} career at \{organization\_1\}. After years of hard work, \{gender\_subj\} became a manager at \{organization\_2\}. Recognized for \{gender\_possessive\_adj\} expertise, \{gender\_subj\} was later recruited as director at \{organization\_3\}. \\
\cmidrule(r){2-3}
                         & Company      & \{subject\} launched \{gender\_possessive\_adj\} first product with support from \{organization\_1\}. \{Gender\_subj\} later collaborated on a major project with \{organization\_2\}. Eventually, \{subject\} was acquired by \{organization\_3\}. \\
\hline
\multirow{8}{*}{Creative Work}   & Person       & \{subject\} discovered a passion for creative work after encountering \{creative\_work\_1\}. In college, \{subject\} analyzed \{creative\_work\_2\} in \{gender\_possessive\_adj\} thesis. Later, \{gender\_subj\}'s award-winning work, inspired by \{creative\_work\_3\}, gained recognition in the creative world. \\
\cmidrule(r){2-3}
                         & Company      & \{subject\} built \{gender\_possessive\_adj\} culture on the influence of \{creative\_work\_1\}. Later, discussions around \{creative\_work\_2\} became common among \{gender\_possessive\_adj\} employees. At a later stage, \{gender\_subj\} added \{creative\_work\_3\} to \{gender\_possessive\_adj\} recommended list for creative development. \\
\bottomrule
\end{tabular}
\end{table*}

%% file: tables/all_entities_synstory_small.tex
\begin{table*}[]
\centering
\resulttablefontsize
\renewcommand{\arraystretch}{1.2}
\setlength{\tabcolsep}{4pt}
\caption{All real-world entities in \syndataSmall.}
\label{table:syndata:small_entities}
\begin{tabular}{llp{0.6\linewidth}}
\toprule
In-Domain / Out-of-Domain   & Real-world Entity Type   & Entity Instances       \\
\hline

\multirow{31}{*}{In-Domain} & \multirow{5}{*}{Person}   &   Martin Luther King Jr., Napoleon Bonaparte, William Wordsworth, William Shakespeare, Genghis Khan, Vincent van Gogh, Mother Teresa, Leonardo da Vinci, Eleanor Roosevelt, Theodore Roosevelt, Albert Einstein, Cleopatra VII, Frida Kahlo, Pablo Picasso, Rosa Parks, Elvis Presley, Joan of Arc, Franklin D. Roosevelt, Marie Antoinette, Henry VIII, Coco Chanel \\
     \cmidrule(r){2-3}
     & \multirow{3}{*}{Language}   &  Polish, Portuguese, English, Hindi, Swedish, German, Spanish, Turkish, Greek, Persian (Farsi), Hebrew, French, Arabic, Gujarati, Bengali, Dutch, Korean, Tamil, Telugu, Italian, Kazakh, Haitian Creole, Punjabi, Swahili \\
     \cmidrule(r){2-3}
     & \multirow{4}{*}{Country}   & Iran, Malaysia, Colombia, Kenya, Armenia, Israel, Maldives, Vietnam, Saudi Arabia, Pakistan, Bangladesh, Turkey, Germany, Czech Republic, United States, Russia, Ukraine, Oman, Japan, South Korea, Belgium, Norway, New Zealand, Indonesia, Denmark, France, India, Spain, Iceland, Greece, Thailand \\
     \cmidrule(r){2-3}
     & \multirow{9}{*}{Event}  & The Reign of Alexander the Great, The Fall of the Berlin Wall, The Spanish Conquest of the Aztecs, The Assassination of Julius Caesar, The Collapse of the Soviet Union, The Battle of Midway, The Surrender of Japan in WWII, Abolition of Slavery in the US, The Establishment of the Ming Dynasty, The Emancipation Proclamation, The Execution of King Louis XVI, The Partition of India and Pakistan, The Assassination of John F. Kennedy, Signing of the Magna Carta, American Civil War, Moon Landing, The Battle of Thermopylae, The Establishment of the People's Republic of China, Fall of Constantinople, The Founding of the United States of America, The Taiping Rebellion, The Vietnam War, The Battle of Waterloo, Civil Rights Movement \\
     \cmidrule(r){2-3}
     & \multirow{5}{*}{Organization}  & Toyota, Human Rights Watch, Sony, Spotify, The Salvation Army, Amazon, Bill \& Melinda Gates Foundation, Apple, The ACLU, Ford, World Food Programme, Amnesty International, Siemens, Johnson \& Johnson, World Health Organization, Nestlé, Alibaba, Airbnb, Walmart \\
     &       &  What primary service or product does \{organization\} provide? \\
     \cmidrule(r){2-3}
     & \multirow{3}{*}{Species}  & pygmy hippo, panda, praying mantis, red-shouldered hawk, swan, humpback whale, crocodile, snow leopard, tiger, king cobra, great horned owl, great white shark, wolverine, bengal tiger, whale shark, bald eagle, wildebeest, harpy eagle\\
     \cmidrule(r){2-3}
     & \multirow{4}{*}{Creative Work}  & The Brothers Karamazov, Oldboy, The Count of Monte Cristo, Jane Eyre, Citizen Kane, The Hobbit, Gangnam Style, A Tale of Two Cities, War and Peace, Goodfellas, The Dark Knight, Brave New World, Catch-22, Pulp Fiction, The Grapes of Wrath\\
\hline
\multirow{11}{*}{Out-of-Domain} & \multirow{1}{*}{Person}  &   Alexander the Great, Machiavelli, Charles Dickens \\
\cmidrule(r){2-3}
     &   \multirow{1}{*}{Language}    &  Afrikaans, Sinhala, Russian, Malay, Ukrainian \\
\cmidrule(r){2-3}
     &   \multirow{1}{*}{Country}    &  Portugal, Italy, Sweden, Netherlands, Poland, Azerbaijan, Hungary \\
\cmidrule(r){2-3}
     &   \multirow{3}{*}{Event}    &  The Boston Tea Party, The Montgomery Bus Boycott, Protestant Reformation, The Haitian Revolution, Napoleonic Wars, French Revolution, The 9/11 Attacks, English Civil War, The Battle of Hastings\\
\cmidrule(r){2-3}
     &   \multirow{1}{*}{Organization}    &  Walt Disney Company \\
\cmidrule(r){2-3}
     &   \multirow{1}{*}{Species}    &  albatross, raccoon, mantis shrimp, giant panda, giraffe, sloth, chameleon \\
\cmidrule(r){2-3}
     &   \multirow{1}{*}{Creative Work}    & Pride and Prejudice, The Road, A Separation, Spirited Away, Pan's Labyrinth \\
\bottomrule
\end{tabular}
\end{table*}

%% file: tables/all_relations_synstory_small.tex
\begin{table*}[]
\centering
\resulttablefontsize
\renewcommand{\arraystretch}{1.2}
\setlength{\tabcolsep}{4pt}
\caption{All relations in \syndataSmall.}
\label{table:syndata:small_relations}
\begin{tabular}{llp{0.6\linewidth}}
\toprule
In-Domain / Out-of-Domain   & Real-world Entity Type   & Relation Template       \\
\hline

\multirow{31}{*}{In-Domain} & \multirow{6}{*}{Person}   &   What occupation is \{person\} most well-known for? \\
     &       &  Where was the birthplace of \{person\}? \\
     &       &  What language was primarily spoken by \{person\}? \\
     &       &  What year did \{person\} pass away? \\
     &       &  What is the religion of \{person\}? \\
     &       &  What year was \{person\} born? \\
     \cmidrule(r){2-3}
     & \multirow{3}{*}{Language}   &   What writing system is used by \{language\}? \\
     &       &  What is the ISO 639‑1 code for \{language\}? \\
     &       &  What region is \{language\} native to? \\
     \cmidrule(r){2-3}
     & \multirow{7}{*}{Country}   &  What is the top-level internet domain for \{country\}? \\
     &       &  What is the currency of \{country\}? \\
     &       &  What is the ISO alpha-2 code for \{country\}? \\
     &       &  Which ethnic group is the largest in \{country\}? \\
     &       &  What is the capital of \{country\}? \\
     &       &  What language in \{country\} has the most speakers? \\
     &       &  What is the calling code for \{country\}? \\
     \cmidrule(r){2-3}
     & \multirow{2}{*}{Event}  & In which country did \{event\} happen? \\
     &       &  Who was the most important leader or figure involved in \{event\}? \\
     \cmidrule(r){2-3}
     & \multirow{5}{*}{Organization}  & Where was \{organization\} established? \\
     &       &  In what year was \{organization\} established? \\
     &       &  Who established \{organization\}? \\
     &       &  What is the primary field or industry of \{organization\}? \\
     &       &  What primary service or product does \{organization\} provide? \\
     \cmidrule(r){2-3}
     & \multirow{3}{*}{Species}  & What is the social structure of \{species\}? \\
     &       &  What is the diet of \{species\}? \\
     &       &  What type of organism is \{species\}? \\
     \cmidrule(r){2-3}
     & \multirow{5}{*}{Creative Work}  & What is the original language of \{creative\_work\}? \\
     &       &  When was \{creative\_work\} released or published? \\
     &       &  Where was \{creative\_work\} produced or created? \\
     &       &  In which country was \{creative\_work\} first released or published? \\
     &       &  What is the genre or style of \{creative\_work\}? \\
\hline
\multirow{11}{*}{Out-of-Domain} & \multirow{1}{*}{Person}   &   $\varnothing$ \\
\cmidrule(r){2-3}
     &   \multirow{1}{*}{Language}    &  What is the name of the alphabet or script of \{language\}? \\
\cmidrule(r){2-3}
     &   \multirow{1}{*}{Country}    &  Which religion has the most followers in \{country\}? \\
\cmidrule(r){2-3}
     &   \multirow{2}{*}{Event}    &  When did \{event\} take place? \\
     &       &  What year did \{event\} end? \\
\cmidrule(r){2-3}
     &   \multirow{1}{*}{Organization}    &  Where is the headquarters of \{organization\} located? \\
\cmidrule(r){2-3}
     &   \multirow{1}{*}{Species}    &  Where is \{species\} primarily native to? \\
\cmidrule(r){2-3}
     &   \multirow{1}{*}{Creative Work}    &  Who is the creator of \{creative\_work\}? \\
\bottomrule
\end{tabular}
\end{table*}

%% file: tables/llama1b_full_result_synstory.tex
\begin{table}[]
\small
\centering
\caption{Main Results on \syndata with \texttt{Llama-3.2-1B-base-QA}. We use the model's LLM-Score on multi-hop questions for efficacy, and the model's LLM-Score on single-hop questions for specificity. OOD (Entity) means using ID relation with OOD entity; OOD (Relation) means using ID entity with OOD relation. \mydagger means the system is out-performed by \catchyname accroding to a paired bootstrapping test ($p=0.05$).}
\label{tab:syndata:main:llama}
\begin{tabular}{l|cc|cc|cc|cc}

\toprule
\multirow{3}{*}{LLM-Score $(\uparrow)$} & \multicolumn{2}{c}{In-Domain} &  \multicolumn{2}{c}{OOD (Entity)}&  \multicolumn{2}{c}{OOD (Rel)}&  \multicolumn{2}{c}{OOD (Both)}\\
& \multicolumn{2}{c}{($2284$)} &  \multicolumn{2}{c}{($1368$)}&  \multicolumn{2}{c}{($421$)}&  \multicolumn{2}{c}{($447$)}\\
& Effi. & Spec. & Effi. & Spec. & Effi. & Spec. & Effi. & Spec. \\
\midrule
                      
\texttt{Llama-3.2-1B-base-QA} & 8.3\mydagger & 94.7\mydagger & 7.1\mydagger
& 94.3 & 8.9\mydagger & 94.2 & 10.9\mydagger & \textbf{90.7} \\
\rowcolor{gray!20} + Prepend &  38.1\mydagger  &  86.2\mydagger & 41.5 & 88.2 & 29.4\mydagger & 82.4 & 31.7 & 79.5 \\ \midrule
+ CPT  {\scriptsize(Full)} & 18.1\mydagger & 80.2\mydagger & 17.0\mydagger & 79.9\mydagger  & 15.6\mydagger & 79.3\mydagger &  12.9\mydagger & 71.1\mydagger \\
+ CPT  {\scriptsize(Mid-Upper)} & 8.5\mydagger  & 93.7\mydagger & 7.6\mydagger & 93.9 & 9.2\mydagger & \textbf{94.3} & 11.5\mydagger & 90.1 \\
+ MEMIT {\scriptsize(\texttt{wikitext-103})} & 12.8\mydagger & 94.4\mydagger & 14.4\mydagger  & 94.4 &  12.0\mydagger & 93.9 & 13.8\mydagger & 90.0 \\
+ MEMIT {\scriptsize(\syndata)} & 12.0\mydagger & 94.6\mydagger & 13.3\mydagger & \textbf{94.5} & 11.1\mydagger & \textbf{94.3} & 11.6\mydagger & 90.2 \\

+ MEND {\scriptsize(with standard config)} & 14.7\mydagger & 89.0\mydagger & 14.2\mydagger & 89.4 & 10.1\mydagger & 91.8  & 10.7\mydagger & 86.3 \\
+ MEND {\scriptsize(Mid-Upper)}  & 12.3\mydagger  & 91.8\mydagger & 11.5\mydagger & 92.9 & 11.5\mydagger  & 92.2 &  12.0\mydagger & 88.1  \\
\midrule
+ \catchyname\ {\scriptsize(Mid-Upper)} & 60.8\mydagger & 91.3\mydagger  & \textbf{36.0}  & 85.4 & 28.4\mydagger & 87.4 & \textbf{18.3} & 84.0  \\ 
+ \catchyname & \textbf{76.7} & \textbf{95.5} & 35.2 & 81.6 & \textbf{34.5} & 84.0 & \textbf{18.3} & 77.5  \\
\bottomrule
\end{tabular}
\end{table}

%% file: tables/llama1b_scalup_synstory.tex
\begin{table}[]
\small
\centering
\addtolength{\tabcolsep}{-2.0pt}
\renewcommand{\arraystretch}{1.2}
\caption{Scale-up experiment of \catchyname on \syndata with \texttt{Llama-3.2-1B-base-QA}. We experiment with more in-domain meta-training instances, and different sizes of hypernetwork by having dedicated hypernetworks per target weight in \texttt{Llama-3.2-1B-base-QA}. We observed that having larger training data and hypernetwork tends to improve performances on Out-of-Domain instances, but it remains challenging.}
\label{tab:syndata:scaleup:llama}
\begin{tabular}{l|cc|cc|cc|cc|cc}
\toprule
  \multirow{2}{*}{LLM-Score $(\uparrow)$} & \multirow{2}{*}{\begin{tabular}{@{}c@{}} Hypernet \\ size (\# Param.) \end{tabular}} & \multirow{2}{*}{\begin{tabular}{@{}c@{}} \# train \\ instances \end{tabular}}  & \multicolumn{2}{c}{\begin{tabular}{@{}c@{}} In-Domain \\ ($2284$) \end{tabular}}  & \multicolumn{2}{c}{\begin{tabular}{@{}c@{}} OOD (Entity) \\ ($1368$) \end{tabular}} & \multicolumn{2}{c}{\begin{tabular}{@{}c@{}} OOD (Relation) \\ ($421$) \end{tabular}} & \multicolumn{2}{c}{\begin{tabular}{@{}c@{}} OOD (Both) \\ ($447$) \end{tabular}} \\

&  &  & Effi. & Spec. & Effi. & Spec. & Effi. & Spec. & Effi. & Spec. \\
\midrule

\multirow{2}{*}{\catchyname}  & \multirow{1}{*}{159M} & 4K & 76.7 & 95.5 & 35.2 & 81.6 & 34.5 & 84.0 & 18.3 & 77.5 \\
\cline{2-11}
& \multirow{1}{*}{2.8B} & 30K & \textbf{97.8} & \textbf{97.1} & \textbf{42.5} & \textbf{87.2} & \textbf{41.8} & \textbf{89.5} & \textbf{20.9} & \textbf{87.8} \\
\bottomrule
\end{tabular}
\vspace{-1em}
\end{table}

%% file: tables/llama1b_ablation_synstory.tex
\begin{table}[]
\small
\centering
\renewcommand{\arraystretch}{1.2}
\caption{Ablation Studies of \catchyname on \syndata with \texttt{Llama-3.2-1B-base-QA}. To reduce compute costs, we run \catchyname\ {\scriptsize (Mid-Upper)}, which targets Layer-\texttt{[10-12]} for editing. ``Upper layer'' is Layer-\texttt{[13-15(top)]}. \mydagger means the system is out-performed by \catchyname\ {\scriptsize (Mid-Upper)} accroding to a paired bootstrapping test ($p=0.05$).}
\label{tab:syndata:ablation:llama}
\begin{tabular}{l|cc|cc|cc|cc}
\toprule
  \multirow{2}{*}{LLM-Score $(\uparrow)$} & \multicolumn{2}{c}{\begin{tabular}{@{}c@{}} In-Domain \\ ($2284$) \end{tabular}}  & \multicolumn{2}{c}{\begin{tabular}{@{}c@{}} OOD (Entity) \\ ($1368$) \end{tabular}} & \multicolumn{2}{c}{\begin{tabular}{@{}c@{}} OOD (Relation) \\ ($421$) \end{tabular}} & \multicolumn{2}{c}{\begin{tabular}{@{}c@{}} OOD (Both) \\ ($447$) \end{tabular}} \\

& Effi. & Spec. & Effi. & Spec. & Effi. & Spec. & Effi. & Spec. \\
\midrule

\catchyname\ {\scriptsize (Mid-Upper)} & \textbf{60.8} & 91.3  & \textbf{36.0}  & 85.4 & \textbf{28.4}  & 87.4 & \textbf{18.3} & 84.0  \\ 

\hspace{0.3cm}propagations $\rightarrow$ paraphrases & 12.4\mydagger  & 91.8 & 10.5\mydagger  & \textbf{93.1} & 11.8\mydagger  & \textbf{93.2} & 12.9\mydagger  & \textbf{89.1} \\

\hspace{0.3cm}all tokens $\rightarrow$ answer tokens & 45.9\mydagger & 91.7 &  34.8 & 89.5 &  20.5\mydagger & 89.7 & 16.2 & 88.3  \\

\hspace{0.3cm}Mid-Upper $\rightarrow$ {Upper} layers  & 42.5\mydagger & \textbf{93.8} & 19.4\mydagger & 84.1 & 20.6\mydagger & 89.1 & 11.5\mydagger & 82.5\\

\bottomrule
\end{tabular}
\vspace{-1em}
\end{table}

%% file: tables/llama1b_efficiency_synstory.tex
\begin{table}[]
\small
\centering
\addtolength{\tabcolsep}{-2.3pt}
\renewcommand{\arraystretch}{1.2}
\caption{Efficiency Evaluation with \texttt{Llama-3.2-1B-base-QA} model on 50 examples. All experiments are run on an NVIDIA RTX A6000 GPU, in a server with an Intel Core i9-10940X CPU@3.30GHz. $^{\text{*}}$: we ran 4 gradient update on the injected fact $\mathbf{f}$, beyond which the drop in loss is marginal (see full hyperparameters in Table~\ref{table:hyper:cpt}).}
\label{tab:efficiency:llama}
\begin{tabular}{l|c|c}

\toprule
           & Max Memory Usage (MiB $\downarrow$) & Total Runtime (Second $\downarrow$) \\
\hline
Base Model & 6059 & 42 \\
\rowcolor{gray!20} + Prepend  & + 28 & + 1 \\
\hline
+ CPT {\scriptsize (Full)$^{\text{*}}$} & + 19132 & + 920 \\
+ MEMIT {\scriptsize(\texttt{wikitext-103})}  & + 4010 & + 1291 \\
+ MEND {\scriptsize(Mid-Upper)}  & + 7550 & + 106 \\
\hline
+ \catchyname\ {\scriptsize(Mid-Upper)} &   + 7542  & + 96 \\
+ \catchyname\  &   + 15163  & + 122 \\
\bottomrule
\end{tabular}
\vspace{-1em}
\end{table}

%% file: 000_main.bbl
\begin{thebibliography}{49}
\providecommand{\natexlab}[1]{#1}
\providecommand{\url}[1]{\texttt{#1}}
\expandafter\ifx\csname urlstyle\endcsname\relax
  \providecommand{\doi}[1]{doi: #1}\else
  \providecommand{\doi}{doi: \begingroup \urlstyle{rm}\Url}\fi

\bibitem[Aky{\"u}rek et~al.(2024)Aky{\"u}rek, Aky{\"u}rek, Choshen, Wijaya, and Andreas]{DCT}
Afra~Feyza Aky{\"u}rek, Ekin Aky{\"u}rek, Leshem Choshen, Derry Wijaya, and Jacob Andreas.
\newblock Deductive closure training of language models for coherence, accuracy, and updatability.
\newblock In Lun-Wei Ku, Andre Martins, and Vivek Srikumar, editors, \emph{Findings of the Association for Computational Linguistics: ACL 2024}, pages 9802--9818, Bangkok, Thailand, August 2024. Association for Computational Linguistics.
\newblock \doi{10.18653/v1/2024.findings-acl.584}.
\newblock URL \url{https://aclanthology.org/2024.findings-acl.584/}.

\bibitem[Berglund et~al.(2024)Berglund, Tong, Kaufmann, Balesni, Stickland, Korbak, and Evans]{reversalCurse}
Lukas Berglund, Meg Tong, Maximilian Kaufmann, Mikita Balesni, Asa~Cooper Stickland, Tomasz Korbak, and Owain Evans.
\newblock The reversal curse: {LLM}s trained on {\textquotedblleft}a is b{\textquotedblright} fail to learn {\textquotedblleft}b is a{\textquotedblright}.
\newblock In \emph{The Twelfth International Conference on Learning Representations}, 2024.
\newblock URL \url{https://openreview.net/forum?id=GPKTIktA0k}.

\bibitem[Chang et~al.(2024)Chang, Park, Ye, Yang, Seo, Chang, and Seo]{minjoon_fact}
Hoyeon Chang, Jinho Park, Seonghyeon Ye, Sohee Yang, Youngkyung Seo, Du-Seong Chang, and Minjoon Seo.
\newblock How do large language models acquire factual knowledge during pretraining?
\newblock In \emph{The Thirty-eighth Annual Conference on Neural Information Processing Systems}, 2024.
\newblock URL \url{https://openreview.net/forum?id=TYdzj1EvBP}.

\bibitem[Chen et~al.(2025)Chen, Geng, Bhaskar, Friedman, and Chen]{REMIX}
Howard Chen, Jiayi Geng, Adithya Bhaskar, Dan Friedman, and Danqi Chen.
\newblock Continual memorization of factoids in language models, 2025.
\newblock URL \url{https://arxiv.org/abs/2411.07175}.

\bibitem[Chen et~al.(2023)Chen, Weiss, Mitchell, Celikyilmaz, and Bosselut]{reckoning}
Zeming Chen, Gail Weiss, Eric Mitchell, Asli Celikyilmaz, and Antoine Bosselut.
\newblock {RECKONING}: Reasoning through dynamic knowledge encoding.
\newblock In \emph{Thirty-seventh Conference on Neural Information Processing Systems}, 2023.
\newblock URL \url{https://openreview.net/forum?id=dUAcAtCuKk}.

\bibitem[Cohen et~al.(2024)Cohen, Biran, Yoran, Globerson, and Geva]{ripple_edit}
Roi Cohen, Eden Biran, Ori Yoran, Amir Globerson, and Mor Geva.
\newblock Evaluating the ripple effects of knowledge editing in language models.
\newblock \emph{Transactions of the Association for Computational Linguistics}, 12:\penalty0 283--298, 2024.
\newblock \doi{10.1162/tacl_a_00644}.
\newblock URL \url{https://aclanthology.org/2024.tacl-1.16/}.

\bibitem[De~Cao et~al.(2021)De~Cao, Aziz, and Titov]{Nicola_De_Cao_21_KE}
Nicola De~Cao, Wilker Aziz, and Ivan Titov.
\newblock {Editing Factual Knowledge in Language Models}.
\newblock In \emph{Proceedings of the Conference on Empirical Methods in Natural Language Processing (EMNLP)}, 2021.

\bibitem[Fang et~al.(2025)Fang, Jiang, Wang, Ma, Shi, Wang, He, and Chua]{alphaedit}
Junfeng Fang, Houcheng Jiang, Kun Wang, Yunshan Ma, Jie Shi, Xiang Wang, Xiangnan He, and Tat-Seng Chua.
\newblock Alphaedit: Null-space constrained model editing for language models.
\newblock In \emph{The Thirteenth International Conference on Learning Representations}, 2025.
\newblock URL \url{https://openreview.net/forum?id=HvSytvg3Jh}.

\bibitem[Franke et~al.(2024)Franke, Hefenbrock, and Hutter]{franke2024preserving}
J{\"o}rg~K.H. Franke, Michael Hefenbrock, and Frank Hutter.
\newblock Preserving principal subspaces to reduce catastrophic forgetting in fine-tuning.
\newblock In \emph{ICLR 2024 Workshop on Mathematical and Empirical Understanding of Foundation Models}, 2024.
\newblock URL \url{https://openreview.net/forum?id=XoWtroECJU}.

\bibitem[Goldman et~al.(2025)Goldman, Shaham, Malkin, Eiger, Hassidim, Matias, Maynez, Gilady, Riesa, Rijhwani, Rimell, Szpektor, Tsarfaty, and Eyal]{ECLeKTic}
Omer Goldman, Uri Shaham, Dan Malkin, Sivan Eiger, Avinatan Hassidim, Yossi Matias, Joshua Maynez, Adi~Mayrav Gilady, Jason Riesa, Shruti Rijhwani, Laura Rimell, Idan Szpektor, Reut Tsarfaty, and Matan Eyal.
\newblock {ECLeKTic: a Novel Challenge Set for Evaluation of Cross-Lingual Knowledge Transfer}, 2025.
\newblock URL \url{https://arxiv.org/abs/2502.21228}.

\bibitem[Gu et~al.(2024)Gu, Xu, Ma, Lu, Ling, Chang, and Peng]{gu-etal-2024-model}
Jia-Chen Gu, Hao-Xiang Xu, Jun-Yu Ma, Pan Lu, Zhen-Hua Ling, Kai-Wei Chang, and Nanyun Peng.
\newblock Model editing harms general abilities of large language models: Regularization to the rescue.
\newblock In Yaser Al-Onaizan, Mohit Bansal, and Yun-Nung Chen, editors, \emph{Proceedings of the 2024 Conference on Empirical Methods in Natural Language Processing}, pages 16801--16819, Miami, Florida, USA, November 2024. Association for Computational Linguistics.
\newblock \doi{10.18653/v1/2024.emnlp-main.934}.
\newblock URL \url{https://aclanthology.org/2024.emnlp-main.934/}.

\bibitem[Gururangan et~al.(2020)Gururangan, Marasovi{\'c}, Swayamdipta, Lo, Beltagy, Downey, and Smith]{Gururangan2020DontSP}
Suchin Gururangan, Ana Marasovi{\'c}, Swabha Swayamdipta, Kyle Lo, Iz~Beltagy, Doug Downey, and Noah~A. Smith.
\newblock Don’t stop pretraining: Adapt language models to domains and tasks.
\newblock \emph{Proceedings of the Annual Meeting of the Association for Computational Linguistics (ACL)}, abs/2004.10964, 2020.

\bibitem[Hartvigsen et~al.(2023)Hartvigsen, Sankaranarayanan, Palangi, Kim, and Ghassemi]{hartvigsen2023aging}
Thomas Hartvigsen, Swami Sankaranarayanan, Hamid Palangi, Yoon Kim, and Marzyeh Ghassemi.
\newblock Aging with {GRACE}: Lifelong model editing with discrete key-value adaptors.
\newblock In \emph{Thirty-seventh Conference on Neural Information Processing Systems}, 2023.
\newblock URL \url{https://openreview.net/forum?id=Oc1SIKxwdV}.

\bibitem[Hase et~al.(2024)Hase, Hofweber, Zhou, Stengel-Eskin, and Bansal]{hase2024fundamental}
Peter Hase, Thomas Hofweber, Xiang Zhou, Elias Stengel-Eskin, and Mohit Bansal.
\newblock Fundamental problems with model editing: How should rational belief revision work in {LLM}s?
\newblock \emph{Transactions on Machine Learning Research}, 2024.
\newblock ISSN 2835-8856.
\newblock URL \url{https://openreview.net/forum?id=LRf19n5Ly3}.

\bibitem[Jiang et~al.(2024)Jiang, Sun, Shi, Rodriguez, Zhou, Neubig, Lin, Yih, and Iyer]{jiang-etal-2024-instruction}
Zhengbao Jiang, Zhiqing Sun, Weijia Shi, Pedro Rodriguez, Chunting Zhou, Graham Neubig, Xi~Lin, Wen-tau Yih, and Srini Iyer.
\newblock Instruction-tuned language models are better knowledge learners.
\newblock In Lun-Wei Ku, Andre Martins, and Vivek Srikumar, editors, \emph{Proceedings of the 62nd Annual Meeting of the Association for Computational Linguistics (Volume 1: Long Papers)}, pages 5421--5434, Bangkok, Thailand, August 2024. Association for Computational Linguistics.
\newblock \doi{10.18653/v1/2024.acl-long.296}.
\newblock URL \url{https://aclanthology.org/2024.acl-long.296/}.

\bibitem[Jin and Ren(2024{\natexlab{a}})]{jin2024demystifying}
Xisen Jin and Xiang Ren.
\newblock Demystifying forgetting in language model fine-tuning with statistical analysis of example associations.
\newblock In \emph{NeurIPS 2024 Workshop on Scalable Continual Learning for Lifelong Foundation Models}, 2024{\natexlab{a}}.
\newblock URL \url{https://openreview.net/forum?id=0d03UdUY0w}.

\bibitem[Jin and Ren(2024{\natexlab{b}})]{jin2024what}
Xisen Jin and Xiang Ren.
\newblock {What Will My Model Forget? Forecasting Forgotten Examples in Language Model Refinement}, 2024{\natexlab{b}}.
\newblock URL \url{https://openreview.net/forum?id=u1eynu9DVf}.

\bibitem[Joshi et~al.(2017)Joshi, Choi, Weld, and Zettlemoyer]{triviaqa}
Mandar Joshi, Eunsol Choi, Daniel Weld, and Luke Zettlemoyer.
\newblock {T}rivia{QA}: A large scale distantly supervised challenge dataset for reading comprehension.
\newblock In Regina Barzilay and Min-Yen Kan, editors, \emph{Proceedings of the 55th Annual Meeting of the Association for Computational Linguistics (Volume 1: Long Papers)}, pages 1601--1611, Vancouver, Canada, July 2017. Association for Computational Linguistics.
\newblock \doi{10.18653/v1/P17-1147}.
\newblock URL \url{https://aclanthology.org/P17-1147/}.

\bibitem[Ke et~al.(2023)Ke, Shao, Lin, Konishi, Kim, and Liu]{ke2023continual}
Zixuan Ke, Yijia Shao, Haowei Lin, Tatsuya Konishi, Gyuhak Kim, and Bing Liu.
\newblock Continual pre-training of language models.
\newblock In \emph{The Eleventh International Conference on Learning Representations}, 2023.
\newblock URL \url{https://openreview.net/forum?id=m_GDIItaI3o}.

\bibitem[Kwiatkowski et~al.(2019)Kwiatkowski, Palomaki, Redfield, Collins, Parikh, Alberti, Epstein, Polosukhin, Devlin, Lee, Toutanova, Jones, Kelcey, Chang, Dai, Uszkoreit, Le, and Petrov]{kwiatkowski2019natural}
Tom Kwiatkowski, Jennimaria Palomaki, Olivia Redfield, Michael Collins, Ankur Parikh, Chris Alberti, Danielle Epstein, Illia Polosukhin, Jacob Devlin, Kenton Lee, Kristina Toutanova, Llion Jones, Matthew Kelcey, Ming-Wei Chang, Andrew~M. Dai, Jakob Uszkoreit, Quoc Le, and Slav Petrov.
\newblock Natural questions: A benchmark for question answering research.
\newblock \emph{Transactions of the Association for Computational Linguistics}, 7:\penalty0 452--466, 2019.
\newblock \doi{10.1162/tacl_a_00276}.
\newblock URL \url{https://aclanthology.org/Q19-1026}.

\bibitem[Levy et~al.(2017)Levy, Seo, Choi, and Zettlemoyer]{zsRE}
Omer Levy, Minjoon Seo, Eunsol Choi, and Luke Zettlemoyer.
\newblock Zero-shot relation extraction via reading comprehension.
\newblock In Roger Levy and Lucia Specia, editors, \emph{Proceedings of the 21st Conference on Computational Natural Language Learning ({C}o{NLL} 2017)}, pages 333--342, Vancouver, Canada, August 2017. Association for Computational Linguistics.
\newblock \doi{10.18653/v1/K17-1034}.
\newblock URL \url{https://aclanthology.org/K17-1034/}.

\bibitem[Li et~al.(2025)Li, Jiang, Chen, Bi, Zhou, Sun, Fang, and Wang]{RLEdit}
Zherui Li, Houcheng Jiang, Hao Chen, Baolong Bi, Zhenhong Zhou, Fei Sun, Junfeng Fang, and Xiang Wang.
\newblock Reinforced lifelong editing for language models, 2025.
\newblock URL \url{https://arxiv.org/abs/2502.05759}.

\bibitem[Liu et~al.(2025)Liu, Pandit, Ye, Choi, and Durrett]{codeupdatearena}
Zeyu~Leo Liu, Shrey Pandit, Xi~Ye, Eunsol Choi, and Greg Durrett.
\newblock {CodeUpdateArena: Benchmarking Knowledge Editing on API Updates}, 2025.
\newblock URL \url{https://arxiv.org/abs/2407.06249}.

\bibitem[Ma et~al.(2024)Ma, Gu, Ling, Liu, and Liu]{bidirection_edit}
Jun-Yu Ma, Jia-Chen Gu, Zhen-Hua Ling, Quan Liu, and Cong Liu.
\newblock Untying the reversal curse via bidirectional language model editing, 2024.
\newblock URL \url{https://arxiv.org/abs/2310.10322}.

\bibitem[Ma et~al.(2025)Ma, Wang, Xu, Ling, and Gu]{ma2025perturbationrestrained}
Jun-Yu Ma, Hong Wang, Hao-Xiang Xu, Zhen-Hua Ling, and Jia-Chen Gu.
\newblock Perturbation-restrained sequential model editing.
\newblock In \emph{The Thirteenth International Conference on Learning Representations}, 2025.
\newblock URL \url{https://openreview.net/forum?id=bfI8cp8qmk}.

\bibitem[Meng et~al.(2022)Meng, Bau, Andonian, and Belinkov]{rome}
Kevin Meng, David Bau, Alex Andonian, and Yonatan Belinkov.
\newblock {Locating and Editing Factual Associations in GPT}.
\newblock In \emph{Proceedings of Advances in Neural Information Processing Systems (NeurIPS)}, 2022.

\bibitem[Meng et~al.(2023)Meng, Sharma, Andonian, Belinkov, and Bau]{memit}
Kevin Meng, Arnab~Sen Sharma, Alex Andonian, Yonatan Belinkov, and David Bau.
\newblock {Mass-Editing Memory in a Transformer}.
\newblock In \emph{International Conference on Learning Representations (ICLR)}, 2023.

\bibitem[Merity et~al.(2017)Merity, Xiong, Bradbury, and Socher]{wiki103}
Stephen Merity, Caiming Xiong, James Bradbury, and Richard Socher.
\newblock Pointer sentinel mixture models.
\newblock In \emph{International Conference on Learning Representations}, 2017.
\newblock URL \url{https://openreview.net/forum?id=Byj72udxe}.

\bibitem[Mitchell et~al.(2022)Mitchell, Lin, Bosselut, Finn, and Manning]{mend}
Eric Mitchell, Charles Lin, Antoine Bosselut, Chelsea Finn, and Christopher~D Manning.
\newblock {Fast Model Editing at Scale}.
\newblock In \emph{International Conference on Learning Representations (ICLR)}, 2022.

\bibitem[Nishi et~al.(2025)Nishi, Okawa, Ramesh, Khona, Tanaka, and Lubana]{nishi2025representation}
Kento Nishi, Maya Okawa, Rahul Ramesh, Mikail Khona, Hidenori Tanaka, and Ekdeep~Singh Lubana.
\newblock Representation shattering in transformers: A synthetic study with knowledge editing, 2025.
\newblock URL \url{https://openreview.net/forum?id=MjFoQAhnl3}.

\bibitem[Onoe et~al.(2022)Onoe, Zhang, Choi, and Durrett]{Onoe2022EntityCB}
Yasumasa Onoe, Michael Zhang, Eunsol Choi, and Greg Durrett.
\newblock Entity cloze by date: What {LM}s know about unseen entities.
\newblock In \emph{Findings of the Association for Computational Linguistics: NAACL 2022}, pages 693--702, Seattle, United States, July 2022. Association for Computational Linguistics.
\newblock URL \url{https://aclanthology.org/2022.findings-naacl.52}.

\bibitem[Onoe et~al.(2023)Onoe, Zhang, Padmanabhan, Durrett, and Choi]{Onoe2023KnowledgeInject}
Yasumasa Onoe, Michael~J.Q. Zhang, Shankar Padmanabhan, Greg Durrett, and Eunsol Choi.
\newblock {Can LMs Learn New Entities from Descriptions? Challenges in Propagating Injected Knowledge}.
\newblock In \emph{Proceedings of the Annual Meeting of the Association for Computational Linguistics (ACL)}, 2023.

\bibitem[Padmanabhan et~al.(2023)Padmanabhan, Onoe, Zhang, Durrett, and Choi]{prop_by_distill}
Shankar Padmanabhan, Yasumasa Onoe, Michael~JQ Zhang, Greg Durrett, and Eunsol Choi.
\newblock Propagating knowledge updates to {LM}s through distillation.
\newblock In \emph{Thirty-seventh Conference on Neural Information Processing Systems}, 2023.
\newblock URL \url{https://openreview.net/forum?id=DFaGf3O7jf}.

\bibitem[Perez et~al.(2018)Perez, Strub, de~Vries, Dumoulin, and Courville]{perez2018film}
Ethan Perez, Florian Strub, Harm de~Vries, Vincent Dumoulin, and Aaron~C. Courville.
\newblock {FiLM: Visual Reasoning with a General Conditioning Layer}.
\newblock In \emph{AAAI}, 2018.

\bibitem[Qin et~al.(2024)Qin, Zhang, Han, Yu, Li, and Ji]{qin-etal-2024-new}
Jiaxin Qin, Zixuan Zhang, Chi Han, Pengfei Yu, Manling Li, and Heng Ji.
\newblock Why does new knowledge create messy ripple effects in {LLM}s?
\newblock In Yaser Al-Onaizan, Mohit Bansal, and Yun-Nung Chen, editors, \emph{Proceedings of the 2024 Conference on Empirical Methods in Natural Language Processing}, pages 12602--12609, Miami, Florida, USA, November 2024. Association for Computational Linguistics.
\newblock \doi{10.18653/v1/2024.emnlp-main.700}.
\newblock URL \url{https://aclanthology.org/2024.emnlp-main.700/}.

\bibitem[Qwen et~al.(2025)Qwen, :, Yang, Yang, Zhang, Hui, Zheng, Yu, Li, Liu, Huang, Wei, Lin, Yang, Tu, Zhang, Yang, Yang, Zhou, Lin, Dang, Lu, Bao, Yang, Yu, Li, Xue, Zhang, Zhu, Men, Lin, Li, Tang, Xia, Ren, Ren, Fan, Su, Zhang, Wan, Liu, Cui, Zhang, and Qiu]{qwen25}
Qwen, :, An~Yang, Baosong Yang, Beichen Zhang, Binyuan Hui, Bo~Zheng, Bowen Yu, Chengyuan Li, Dayiheng Liu, Fei Huang, Haoran Wei, Huan Lin, Jian Yang, Jianhong Tu, Jianwei Zhang, Jianxin Yang, Jiaxi Yang, Jingren Zhou, Junyang Lin, Kai Dang, Keming Lu, Keqin Bao, Kexin Yang, Le~Yu, Mei Li, Mingfeng Xue, Pei Zhang, Qin Zhu, Rui Men, Runji Lin, Tianhao Li, Tianyi Tang, Tingyu Xia, Xingzhang Ren, Xuancheng Ren, Yang Fan, Yang Su, Yichang Zhang, Yu~Wan, Yuqiong Liu, Zeyu Cui, Zhenru Zhang, and Zihan Qiu.
\newblock Qwen2.5 technical report, 2025.
\newblock URL \url{https://arxiv.org/abs/2412.15115}.

\bibitem[Scialanga et~al.(2025)Scialanga, Laugel, Grari, and Detyniecki]{sake}
Marco Scialanga, Thibault Laugel, Vincent Grari, and Marcin Detyniecki.
\newblock {SAKE: Steering Activations for Knowledge Editing}, 2025.
\newblock URL \url{https://arxiv.org/abs/2503.01751}.

\bibitem[Sinitsin et~al.(2020)Sinitsin, Plokhotnyuk, Pyrkin, Popov, and Babenko]{Sinitsin2020Editable}
Anton Sinitsin, Vsevolod Plokhotnyuk, Dmitry Pyrkin, Sergei Popov, and Artem Babenko.
\newblock Editable neural networks.
\newblock In \emph{International Conference on Learning Representations}, 2020.
\newblock URL \url{https://openreview.net/forum?id=HJedXaEtvS}.

\bibitem[Tan et~al.(2024)Tan, Zhang, and Fu]{malmen_edit}
Chenmien Tan, Ge~Zhang, and Jie Fu.
\newblock Massive editing for large language models via meta learning.
\newblock In \emph{ICLR}, 2024.
\newblock URL \url{https://openreview.net/forum?id=L6L1CJQ2PE}.

\bibitem[Trivedi et~al.(2022)Trivedi, Balasubramanian, Khot, and Sabharwal]{trivedi-etal-2022-musique}
Harsh Trivedi, Niranjan Balasubramanian, Tushar Khot, and Ashish Sabharwal.
\newblock {M}u{S}i{Q}ue: Multihop questions via single-hop question composition.
\newblock \emph{Transactions of the Association for Computational Linguistics}, 10:\penalty0 539--554, 2022.
\newblock \doi{10.1162/tacl_a_00475}.
\newblock URL \url{https://aclanthology.org/2022.tacl-1.31/}.

\bibitem[Wang et~al.(2024)Wang, Zhang, Tian, Xi, Yao, Xu, Wang, Mao, Wang, Cheng, Liu, Ni, Zheng, and Chen]{wang2023easyedit}
Peng Wang, Ningyu Zhang, Bozhong Tian, Zekun Xi, Yunzhi Yao, Ziwen Xu, Mengru Wang, Shengyu Mao, Xiaohan Wang, Siyuan Cheng, Kangwei Liu, Yuansheng Ni, Guozhou Zheng, and Huajun Chen.
\newblock {E}asy{E}dit: An easy-to-use knowledge editing framework for large language models.
\newblock In Yixin Cao, Yang Feng, and Deyi Xiong, editors, \emph{Proceedings of the 62nd Annual Meeting of the Association for Computational Linguistics (Volume 3: System Demonstrations)}, pages 82--93, Bangkok, Thailand, August 2024. Association for Computational Linguistics.
\newblock \doi{10.18653/v1/2024.acl-demos.9}.
\newblock URL \url{https://aclanthology.org/2024.acl-demos.9/}.

\bibitem[Xu et~al.(2025)Xu, Ji, Cao, Lu, Lin, Han, He, Sun, Li, and Sun]{deepknowledge}
Ruoxi Xu, Yunjie Ji, Boxi Cao, Yaojie Lu, Hongyu Lin, Xianpei Han, Ben He, Yingfei Sun, Xiangang Li, and Le~Sun.
\newblock Memorizing is not enough: Deep knowledge injection through reasoning, 2025.
\newblock URL \url{https://arxiv.org/abs/2504.00472}.

\bibitem[Yang et~al.(2018)Yang, Qi, Zhang, Bengio, Cohen, Salakhutdinov, and Manning]{yang-etal-2018-hotpotqa}
Zhilin Yang, Peng Qi, Saizheng Zhang, Yoshua Bengio, William Cohen, Ruslan Salakhutdinov, and Christopher~D. Manning.
\newblock {H}otpot{QA}: A dataset for diverse, explainable multi-hop question answering.
\newblock In Ellen Riloff, David Chiang, Julia Hockenmaier, and Jun{'}ichi Tsujii, editors, \emph{Proceedings of the 2018 Conference on Empirical Methods in Natural Language Processing}, pages 2369--2380, Brussels, Belgium, October-November 2018. Association for Computational Linguistics.
\newblock \doi{10.18653/v1/D18-1259}.
\newblock URL \url{https://aclanthology.org/D18-1259/}.

\bibitem[Yang et~al.(2024)Yang, Band, Li, Candès, and Hashimoto]{synCPT}
Zitong Yang, Neil Band, Shuangping Li, Emmanuel Candès, and Tatsunori Hashimoto.
\newblock Synthetic continued pretraining, 2024.
\newblock URL \url{https://arxiv.org/abs/2409.07431}.

\bibitem[Yao et~al.(2025)Yao, Fang, Gu, Zhang, Deng, Chen, and Peng]{cake}
Yunzhi Yao, Jizhan Fang, Jia-Chen Gu, Ningyu Zhang, Shumin Deng, Huajun Chen, and Nanyun Peng.
\newblock Cake: Circuit-aware editing enables generalizable knowledge learners, 2025.
\newblock URL \url{https://arxiv.org/abs/2503.16356}.

\bibitem[Zhang et~al.(2024)Zhang, Chen, Li, Wang, He, Huang, Xue{'}, and Huang]{zhang-etal-2024-dafnet}
Taolin Zhang, Qizhou Chen, Dongyang Li, Chengyu Wang, Xiaofeng He, Longtao Huang, Hui Xue{'}, and Jun Huang.
\newblock {DAFN}et: Dynamic auxiliary fusion for sequential model editing in large language models.
\newblock In Lun-Wei Ku, Andre Martins, and Vivek Srikumar, editors, \emph{Findings of the Association for Computational Linguistics: ACL 2024}, pages 1588--1602, Bangkok, Thailand, August 2024. Association for Computational Linguistics.
\newblock \doi{10.18653/v1/2024.findings-acl.92}.
\newblock URL \url{https://aclanthology.org/2024.findings-acl.92/}.

\bibitem[Zheng et~al.(2023)Zheng, Li, Dong, Fan, Wu, Xu, and Chang]{IKE}
Ce~Zheng, Lei Li, Qingxiu Dong, Yuxuan Fan, Zhiyong Wu, Jingjing Xu, and Baobao Chang.
\newblock Can we edit factual knowledge by in-context learning?
\newblock In \emph{The 2023 Conference on Empirical Methods in Natural Language Processing}, 2023.
\newblock URL \url{https://openreview.net/forum?id=hsjQHAM8MV}.

\bibitem[Zhong et~al.(2025)Zhong, Lu, Shao, Bhushanam, Du, Wan, Shi, Zha, Wang, Liu, Zhou, Xu, Chang, Feng, Chaudhary, and Hu]{zhong2025mquakeremastered}
Shaochen Zhong, Yifan Lu, Lize Shao, Bhargav Bhushanam, Xiaocong Du, Yixin Wan, Yucheng Shi, Daochen Zha, Yiwei Wang, Ninghao Liu, Kaixiong Zhou, Shuai Xu, Kai-Wei Chang, Louis Feng, Vipin Chaudhary, and Xia Hu.
\newblock {MQ}u{AKE}-remastered: Multi-hop knowledge editing can only be advanced with reliable evaluations.
\newblock In \emph{The Thirteenth International Conference on Learning Representations}, 2025.
\newblock URL \url{https://openreview.net/forum?id=m9wG6ai2Xk}.

\bibitem[Zhong et~al.(2023)Zhong, Wu, Manning, Potts, and Chen]{zhong2023mquake}
Zexuan Zhong, Zhengxuan Wu, Christopher~D Manning, Christopher Potts, and Danqi Chen.
\newblock {MQuAKE}: Assessing knowledge editing in language models via multi-hop questions.
\newblock \emph{arXiv preprint arXiv:2305.14795}, 2023.

\end{thebibliography}
